\documentclass[11pt]{article}
\usepackage[final]{acl}

\usepackage{times}
\usepackage{latexsym}

\usepackage[T1]{fontenc}
\usepackage[utf8]{inputenc}

\usepackage{microtype}

\usepackage{inconsolata}

\usepackage{graphicx}
\usepackage{amsmath}
\usepackage{amssymb}
\usepackage{amsfonts}
\usepackage{booktabs} 
\usepackage{multirow}
\usepackage{bm}
\usepackage{booktabs}
\usepackage{multirow}
\usepackage{enumitem}
\usepackage[most]{tcolorbox}
\usepackage{hyperref}
\usepackage[dvipsnames]{xcolor}
\usepackage{makecell}

\newcommand{\greenup}{\textcolor{ForestGreen}{\uparrow}}
\newcommand{\reddown}{\textcolor{Red}{\downarrow}}

\title{Single-Pass, Depth-Selective Reading for \\ Multi-Aspect Sentiment Analysis}

\author{
Yan Xia\textsuperscript{1,2}\thanks{equal contribution; authors are listed alphabetically by first name.}\,\,
Zhuangzhuang Pan\textsuperscript{1}\footnotemark[1]\,\,
Amirrudin Kamsin\textsuperscript{1}\,\,
Chee Seng Chan\textsuperscript{1,3}\thanks{Corresponding author (cs.chan@um.edu.my).} \\
\textsuperscript{1}Universiti Malaya, Malaysia \\
\textsuperscript{2}Suzhou University of Technology, China \\
\textsuperscript{3}VinUniversity, Vietnam \\
\texttt{\{23072126; 23078403\}@siswa.um.edu.my, {\{amir; cs.chan\}}@um.edu.my}
}

\begin{document}
\maketitle
\begin{abstract}
Aspect-Term Sentiment Analysis (ATSA) in multi-aspect sentences faces a fundamental tradeoff between efficiency and expressiveness. Existing models either re-encode the sentence for each aspect or rely on static use of deep representations, leading to redundant computation and limited adaptivity. We argue that Transformer depth is a costly, queryable resource, and propose \textbf{DABS}, a single-pass inference framework that encodes each sentence once to construct a reusable, depth-ordered substrate. Each aspect then queries this shared representation to selectively read relevant tokens and abstraction levels, without re-encoding. This decouples shared sentence encoding from lightweight, aspect-conditioned readout. Experiments on four ATSA benchmarks show that DABS achieves competitive performance while reducing end-to-end computation by up to 60\% in multi-aspect settings ($M \ge 2$). Further analyses indicate that adaptive depth querying is most beneficial for linguistically complex cases such as negation and contrast. Code is publicly available at \url{https://github.com/panzhzh/acl-dabs}.
\end{abstract}

\section{Introduction}
\label{sec:intro}

Aspect-Term Sentiment Analysis (ATSA) predicts the polarity of a given aspect term within a sentence ~\cite{zhang-etal-2023-span,wang-etal-2023-supervised}. Although pretrained Transformer-based systems are strong ~\cite{ma-etal-2023-amr,cabello-akujuobi-2024-simple,ding-etal-2024-boosting}, multi-aspect sentences expose a mismatch between how the task is structured and how computation is typically performed ~\cite{seo-etal-2024-make,jin-etal-2025-aspect}. In a sentence like “The \textit{food} was excellent but the \textit{service} was terrible, and the \textit{atmosphere} felt cramped”, the aspects behave like parallel queries over largely shared context. Yet, many existing methods either (i) process each aspect separately, incurring linear cost in the number of aspects ~\cite{zheng2024you}, or (ii) collapse representations in a uniform way that can dilute aspect-specific evidence ~\cite{wagner-foster-2023-investigating,zhu-etal-2024-pinpointing}. Consequently, researchers face a binary choice between the computational overhead of per-aspect re-encoding and the information loss inherent in static, depth-compressed representations ~\cite{bao-etal-2023-opinion,lv-etal-2023-efficient}.

We argue that multi-aspect ATSA is better viewed as aspect-conditioned querying over a shared sentence representation, where each aspect may benefit from a different amount of representational depth. Deeper representations are often helpful for compositional phenomena such as negation, contrast, and discourse reversal, while other aspects can be resolved using shallower or intermediate evidence ~\cite{chai-etal-2023-aspect,petty-etal-2024-impact,xu-etal-2024-iacos}. Treating all aspects as equally ``\textit{deep}'' can therefore obscure how linguistic difficulty relates to abstraction level. This motivates viewing Transformer depth as a \textit{queryable resource}, rather than as a uniform feature pool ~\cite{bae-etal-2023-fast,elhoushi-etal-2024-layerskip}.

Under this view, each aspect navigates a trade-off between efficiency and semantic richness. In this framing, \textit{deeper layers provide more nuanced information, but the model should limit its depth of traversal to the specific requirements of the aspect.} 

To this end, we propose \textbf{DABS} (\textbf{D}epth-Ordered \textbf{A}ggregation and \textbf{B}udget-Aware \textbf{S}election), a \textit{single-pass} framework that separates heavy sentence-level encoding from lightweight, aspect-specific reads. DABS first constructs a reusable \textit{depth substrate} that functions as an aspect-agnostic bank of representations exposing multiple abstraction levels from a single encoder pass. At inference time, each aspect adaptively queries the shared substrate to resolve both \emph{where} relevant evidence lies in the sentence and \emph{how much abstraction} is required to interpret it, without re-encoding the input.

Together, this formulation enables budget-aware selection, allowing representational capacity across depth to be allocated adaptively while amortizing the encoder computation. Our subsequent analysis, conducted via controlled depth masking, demonstrates that prediction quality is sensitive to specific depth regions, particularly in linguistically challenging cases, such as negation and contrast.

Overall, our contributions are:
\begin{itemize}%[itemsep=0pt, parsep=0pt, topsep=0pt, partopsep=0pt]
\item We cast multi-aspect ATSA as aspect-conditioned queries over a shared depth substrate, explicitly linking marginal compute to aspect-specific expressiveness.
\item We propose DABS, a single-pass architecture that constructs a reusable depth substrate per sentence, enabling aspects to query both evidence location (tokens) and abstraction level (depth) without re-encoding the input.
\item We provide controlled depth-masking analyses demonstrating that performance depends non-uniformly on specific depth regions, supporting the view that depth selection plays a functional role in sentiment reasoning rather than serving as a redundant feature ensemble.
\item Across four benchmarks, DABS achieves competitive results. In multi-aspect settings, it reduces end-to-end computation by up to 60\%\footnote{For single-aspect sentences ($M=1$), DABS introduces a fixed overhead due to depth-substrate construction. This cost is paid only once per sentence and can be significantly amortized for $M \ge 2$. See Figure~\ref{fig:multi_aspect} and Appendix~\ref{app:experiment_reproducibility} for detailed efficiency breakdowns.} compared to standard per-aspect encoding baselines, showing that explicit depth allocation can improve efficiency while preserving fine-grained sentiment modeling.
\end{itemize}

\section{Related Work} 
\label{sec:related}
\paragraph{Structure-aware ATSA.} A substantial body of work injects syntactic structure (typically dependency graphs) to align aspects with opinion expressions and to route information along aspect-centered paths ~\cite{bao-etal-2023-exploring,yin2024textgt,zhang-etal-2023-aspect}. While these models offer explicit alignment signals, they rely on external parsers that are brittle on noisy text and computationally expensive to run. Furthermore, many structure-aware designs fundamentally rely on aspect-specific graph construction (\textit{\textit{e.g.}}, graph reweighting or masking), which prevents representation sharing and incurs linear computational costs in multi-aspect settings.

\paragraph{PLM Fine-Tuning and Hybrid Models.} The dominant paradigm fine-tunes pretrained encoders by concatenating aspect-sentence pairs or injecting aspect markers ~\cite{zheng-etal-2024-instruction,gou-etal-2023-mvp,mukherjee-etal-2023-contraste}. While effective, this approach suffers from two limitations central to our study. First, it treats the sentence as an aspect-dependent object, necessitating a full encoder pass for every aspect \citep{zhang-etal-2023-span,zheng2024you}. Second, regarding representational depth, these models typically adopt a static usage pattern, either relying solely on the final layer or performing fixed scalar mixing across all layers \citep{he-etal-2025-cabilstm,ruan-etal-2025-layalign}. Some works have attempted to complement PLMs with local mechanisms (\textit{e.g.}, CNNs or RNNs) to capture compositional cues ~\cite{feng-etal-2023-aspect,wang-etal-2024-dagcn}. However, these are invariably implemented as static architectural layers within a per-aspect pipeline. Unlike DABS, they do not decouple the shared representation from the query, nor do they treat depth as a dynamically allocatable resource.

\paragraph{Generative and LLM-based ATSA.} Recent approaches cast ATSA as instruction-following generation or few-shot in-context learning ~\cite{scaria-etal-2024-instructabsa,shen-etal-2025-zero,hellwig-etal-2025-exploring,zheng-etal-2024-instruction}. While Large Language Models (LLMs) encode rich reasoning patterns, they introduce significant inference latency and lack the fine-grained controllability required for span-level polarity classification. This makes them less aligned with the high-throughput, resource-constrained inference goals of this work \citep{paul-etal-2023-large,bodke-etal-2025-pastel}.

\begin{figure*}[t] 
	\centering 
	\includegraphics[width=\linewidth, keepaspectratio]{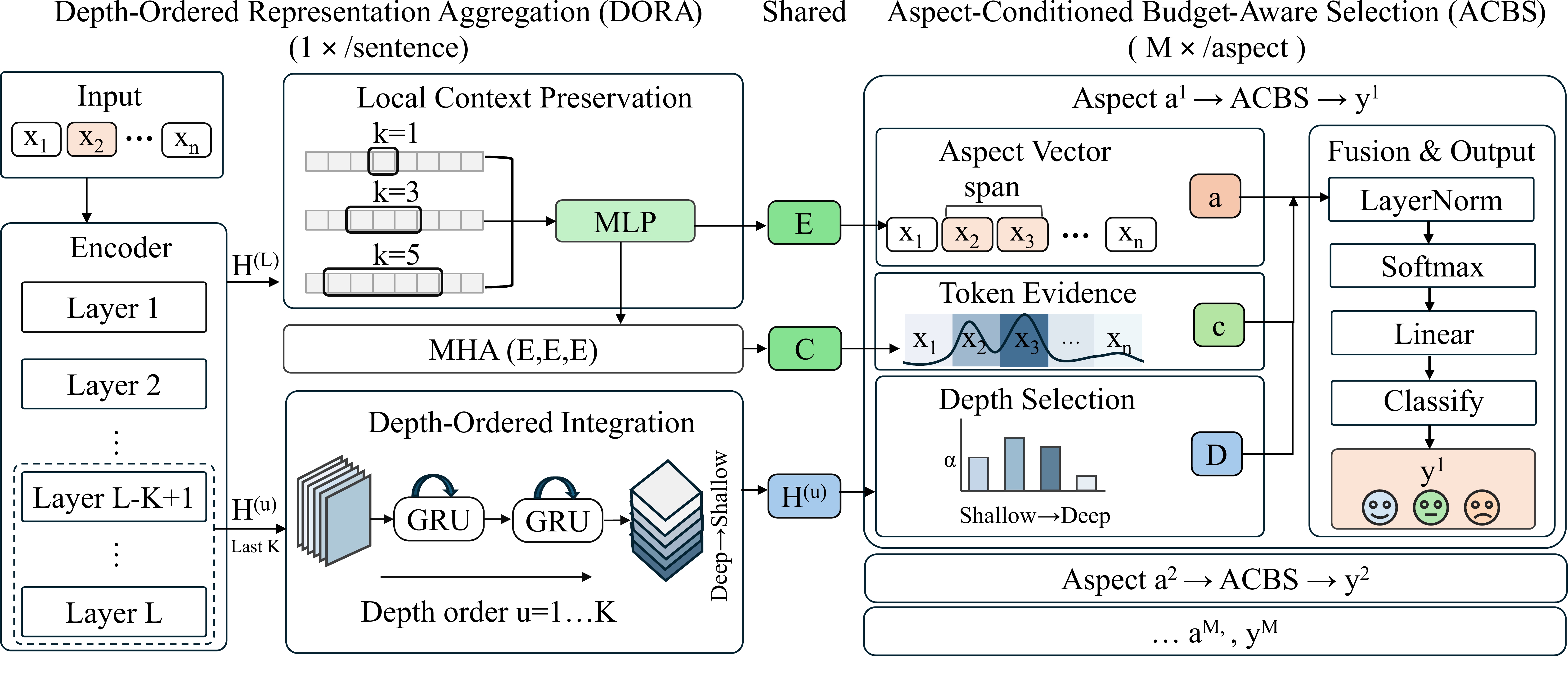} 
	\caption{Overview of the proposed DABS framework. DORA constructs a shared depth substrate via a single encoder pass, and ACBS performs aspect-conditioned token localization and budget-aware depth selection.}
	\label{fig:overview} 
    \vspace{-2mm}
\end{figure*}

\section{Methods}
\label{sec:methods}

\subsection{Problem Definition}
\label{subsec:problem}

ATSA focuses on predicting sentiment polarities for specified aspect terms. Formally, we consider a tokenized sentence $\mathbf{x}=(x_1,\cdots,x_n)$ together with a set of annotated target aspects $\mathcal{A}=\{a^{(k)}\}_{k=1}^{M}$. Each aspect $a^{(k)}$ is represented by its position interval $[i_k,j_k]$, where $1\le i_k\le j_k\le n$. Our task is to model the conditional distribution $p_\theta\!\left(y^{(k)} \mid \mathbf{x},\, a^{(k)}\right)$, where $y^{(k)} \in \{\mathrm{positive}, \mathrm{neutral}, \mathrm{negative}\}$ and $\theta$ denotes the parameters of both the pretrained encoder and the task-specific components.

Unlike standard approaches that re-encode $\mathbf{x}$ for every $a^{(k)}$, we formulate this as a \textbf{query-retrieval problem}: \textit{the sentence acts as a shared depth substrate, and each aspect is a budget-aware query that must independently localize evidence and select the necessary representational depth}.

\subsection{Overview of Budget-Aware Inference}
\label{subsec:overview}

We view multi-aspect ATSA as a \emph{query problem over shared representations}, rather than as a collection of independent classification instances. In a multi-aspect sentence, all aspects observe the same underlying context, but differ in how much \emph{representational depth} they require to resolve their sentiment. Some aspects can be decided from shallow lexical cues, while others require deeper compositional reasoning involving negation, contrast, or long-range dependency.

Existing approaches do not expose this distinction. They either re-encode the sentence for each aspect, implicitly assuming that every aspect requires full depth, or collapse all layers into a static vector, removing the model’s ability to adapt depth usage. As a result, Transformer depth is treated as a free and uniform feature pool, rather than as a constrained and ordered resource.

We propose \textbf{DABS}, a single-pass inference framework that makes Transformer depth explicitly queryable under fixed encoder computation. DABS decomposes inference into two lightweight stages: 
(i) \textbf{Depth-Ordered Representation Aggregation (DORA)}, which constructs a reusable, depth-ordered substrate from a single encoder pass, and (ii) \textbf{Aspect-Conditioned Budget-Aware Selection (ACBS)}, which performs aspect-specific readout over this substrate. Both stages operate as readout mechanisms over existing representations, rather than adding new iterative reasoning or dynamic depth traversal. Figure~\ref{fig:overview} summarizes the overall architecture.

\subsection{Stage I: Constructing a Shared Depth-Ordered Substrate}
\label{subsec:dora}

We first construct a reusable depth substrate via 
\textbf{Depth-Ordered Representation Aggregation (DORA)} that preserves both local compositionality and the multi-level nature of Transformer depth. This aspect-agnostic module constructs the shared depth substrate by reshaping hierarchical encoder hidden states $\{H^{(\ell)}\}_{\ell=1}^{L}$ via two paths.

\subsubsection{Preserving Local Sentiment Cues}
While Transformers capture long-range dependencies, global mixing can weaken short-range cues such as negation markers adjacent to sentiment expressions (\textit{e.g.}, ``\textit{not good}''). To keep these local patterns explicit in the shared substrate, we apply a lightweight local refinement on the final encoder states $H^{(L)}$ using depthwise-separable filters with small kernel sizes $k\in\{1,3,5\}$:
\begin{equation}
\tilde E
= \mathrm{Concat}\big(\{\mathrm{Conv}_k(H^{(L)})\}_{k\in\{1,3,5\}}\big),
\end{equation}
\begin{equation}
E
= \mathrm{LayerNorm}\!\big(\tilde E\,\mathbf W_c + H^{(L)}\big).
\end{equation}
Here, $k\in\{1,3,5\}$ provides a small set of receptive fields with minimal overhead.

\subsubsection{Integrating Representations Across Depth}
Crucially, we must expose the evolution of representations across layers without treating layers as exchangeable. We therefore apply a gated recurrence over the last $K$ layers to form a depth-ordered stack $\{\tilde H^{(u)}\}_{u=1}^{K}$. For each token position $t$, the recurrence state is initialized at the first depth step and updated in increasing depth order:
\begin{align}
    \mathbf{s}_{1,t} &= H^{(L-K+1)}_t, \\
    \begin{split}
        \mathbf{s}_{u,t} &= \mathrm{GRUCell}\!\big(H^{(L-K+u)}_t,\, \mathbf{s}_{u-1,t}\big), \\
        &\qquad\qquad\quad u=2,\ldots,K.
    \end{split}
    \label{eq:gru-rec}
\end{align}
We then form the depth-ordered outputs by residual addition and LayerNorm:
\begin{align}
    \begin{split}
        \tilde H^{(1)} &= \mathrm{LayerNorm}\!\big(\beta\,\mathbf{s}_{1} + H^{(L-K+1)}\big), \\
        &\qquad\qquad\quad \beta\in\mathbb{R},
    \end{split} \\
    \begin{split}
        \tilde H^{(u)} &= \mathrm{LayerNorm}\!\big(\mathbf{s}_{u} + H^{(L-K+u)}\big), \\
        &\qquad\qquad\quad u=2,\ldots,K.
    \end{split}
    \label{eq:gru-out}
\end{align}
We implement this module as \textbf{DepthGRU} whereby a GRUCell is applied over the last $K$ layers in increasing depth order, producing a reusable stack $\{\tilde H^{(u)}\}_{u=1}^{K}$ for querying. Note that DepthGRU is one concrete instantiation of ordered depth integration, whereby any mechanism that preserves monotonic abstraction flow could serve the same role.

\subsection{Stage II: Aspect-Conditioned Reading from the Substrate}
\label{subsec:acbs}

Given the shared substrate, each aspect performs a lightweight read via \textbf{Aspect-Conditioned Budget-Aware Selection (ACBS)}. ACBS performs a two-axis read over the shared depth substrate to determine \textbf{\emph{where}} to read (token evidence) and \textbf{\emph{how deep}} to read (depth preference).

\paragraph{Aspect vector.}
For an aspect $a^{(k)}=[i_k,j_k]$, we compute an aspect vector by averaging the enhanced representations within the span, $\mathbf{a}^{(k)}=\frac{1}{j_k-i_k+1}\sum_{t=i_k}^{j_k}E_t$.

\paragraph{Shared context reorganization.}
We compute a reusable contextualized representation once per sentence by applying multi-head attention over $E$, as
$C=\mathrm{MHA}(E,E,E).$

\subsubsection{Axis 1: Selecting Relevant Evidence Tokens}
Standard token-level Softmax attention imposes competition between tokens. To allow evidence to be accumulated across multiple positions, we employ independent sigmoid gating:
\begin{equation}
w_t = \sigma\!\big(\mathrm{MLP}([C_t;\mathbf{a}])\big), \quad
\mathbf{c} = \frac{\sum_{t=1}^{n} w_t\, C_t}{\sum_{t=1}^{n} w_t + \varepsilon}.
\label{eq:token_select}
\end{equation}

\subsubsection{Axis 2: Selecting an Appropriate Depth Level}
We predict an aspect-conditioned distribution over the depth substrate:
\begin{equation}
    \bm{\alpha}=\mathrm{Softmax}\!\Big(\tfrac{1}{\tau_\alpha}\,\mathrm{MLP}([\mathbf{a};\mathrm{pool}(C)])\Big).
    \label{eq:alpha}
\end{equation}
where $\sum_{u=1}^{K}\alpha_u=1$, $\mathrm{pool}(C)=\frac{1}{n}\sum_{t=1}^{n} C_t$ denotes mean pooling over the sequence. The corresponding depth summary is
$\mathcal{D} = \frac{1}{n}\sum_{t=1}^{n}\sum_{u=1}^{K}\alpha_u\,\tilde H_t^{(u)}.$

\subsubsection{Combining Evidence and Producing Predictions}
\label{subsubsec:gated_fusion}
Token evidence vector $\mathbf{c}$, depth summary $\mathcal{D}$, and aspect vector $\mathbf{a}$ capture complementary signals. We first normalize them, then compute fusion gates:
\begin{equation}
    \begin{split}
        \hat{\mathbf{x}} &= \mathrm{LayerNorm}(\mathbf{x}), \quad \forall \mathbf{x} \in \{\mathbf{c}, \mathcal{D}, \mathbf{a}\}, \\
        \mathbf{g} &= \mathrm{Softmax}\!\Big(\tfrac{1}{\tau_g}\,\mathrm{MLP}([\hat{\mathbf{c}};\hat{\mathcal{D}};\hat{\mathbf{a}}])\Big), \\
        \mathbf{h} &= g_1\hat{\mathbf{c}} + g_2\hat{\mathcal{D}} + g_3\hat{\mathbf{a}}.
    \end{split}
    \label{eq:fusion}
\end{equation}
We obtain logits $\mathbf{z}\leftarrow \mathrm{Linear}(\mathbf{h})$ and
$p_\theta\leftarrow \mathrm{Softmax}(\mathbf{z})$.

\subsection{Training Objectives}
\label{subsec:algo}

We train end-to-end using a cross-entropy classification loss $\mathcal{L}_{\mathrm{cls}}$. We further apply three regularizers (described next) to stabilize selection and discourage degenerate solutions:

\subsubsection{Sparsity Regularization ($\mathcal{R}_{\text{sparse}}$).}
To encourage the model to focus only on the most informative sentiment cues and filter out irrelevant background noise, we impose a sparsity penalty. This regularizer minimizes the average activation of the token selection gates: $\mathcal{R}_{\text{sparse}} = \frac{1}{n} \sum_{t=1}^{n} w_t,$
where $n$ is the sequence length and $w_t \in [0,1]$ represents the learned importance weight for the $t$-th token. By minimizing $\mathcal{R}_{\text{sparse}}$, we force the model to be selective, preventing the trivial solution where all tokens are retained (\textit{i.e.}, $w_t \approx 1$).

\subsubsection{Span masking ($\mathcal{R}_{\text{mask}}$).}
We discourage the token selector from relying on tokens inside the aspect span itself.
Let $\bm{m}\in\{0,1\}^n$ be a binary mask such that $m_t=1$ iff $i_k\le t\le j_k$.
We penalize gate activations on the span, 
$\mathcal{R}_{\text{mask}}=\mathrm{BCE}(w\odot \bm{m},\mathbf{0}),$
where $\mathrm{BCE}(\cdot,\cdot)$ denotes elementwise binary cross-entropy averaged over positions. Equivalently, $\|w\odot \bm{m}\|_1$ yields the same effect, while remaining neutral outside the span.

\subsubsection{Fusion-gate Entropy ($\mathcal{R}_{\text{gate}}$).}
To discourage premature collapse of the fusion module onto a single information source (Context, Layer, or Aspect), we maximize the entropy of the gating distribution,
$\mathcal{R}_{\text{gate}} = \sum_{i=1}^3 g_i \log g_i,$
where $g_i$ corresponds to the normalized gating weight for the $i$-th branch (context, layer, and aspect, respectively). Minimizing $\mathcal{R}_{\text{gate}}$ penalizes low-entropy (overly peaked) distributions, thereby discouraging premature collapse onto a single branch.

As a summary, the total objective is:
	\begin{equation}
	\mathcal{L}_{\text{total}} = \mathcal{L}_{\mathrm{cls}}
	+ \lambda_s \mathcal{R}_{\text{sparse}}
	+ \lambda_m \mathcal{R}_{\text{mask}}
	+ \lambda_{\text{ent}} \mathcal{R}_{\text{gate}}.
	\label{eq:total-loss}
	\end{equation}

\subsection{Computational Amortization}
The efficiency of DABS stems from decoupling the shared depth substrate from per-aspect queries. In sentences with $M$ aspects, the complexity scales:
\begin{equation}
\begin{aligned}
\text{Cost}
&\approx
\underbrace{\mathcal{O}\!\left(\text{Encoder} + \text{DORA}\right)}_{\text{Fixed Cost (Paid Once)}} \\
&\quad +\;
M \cdot \underbrace{\mathcal{O}\!\left(\text{Selection}\right)}_{\text{Lightweight per-aspect cost}}.
\end{aligned}
\label{eq:complexity}
\end{equation}
This amortization is most relevant when $M>1$ or when compute is constrained, where repeated re-encoding of the full encoder per-aspect is costly.

%%%%%%%%%%%%%%%%%%%%%%%%%%%%%%%%%%%%%%%%%%%%
\section{Experiment}
\label{sec:experiments}

\subsection{Setup and Baselines}
\label{subsec:setup}
We evaluate DABS on four standard ATSA benchmarks: SemEval-2014 Laptop (Lap14) and Restaurant (Rest14), SemEval-2015 Restaurant (Rest15), and SemEval-2016 Restaurant (Rest16) ~\cite{pontiki-etal-2014-semeval,pontiki-etal-2015-semeval,pontiki-etal-2016-semeval}. Detailed statistics are provided in Appendix~\ref{app:dataset_stats}. We report Accuracy (Acc) and Macro-F1 (MF1). We follow the official SemEval train/test protocol and report results on the official test split.

\paragraph{Aspect multiplicity statistics.}
Since the efficiency gain of DABS is realized through amortization across aspect queries, we report sentence-level aspect multiplicity on the official test splits in Table~\ref{tab:aspect_multiplicity}. This contextualizes the practical relevance of the multi-aspect setting targeted by DABS on standard ATSA benchmarks.

\begin{table}[t]
    \small
    \centering
    \caption{Sentence-level aspect multiplicity statistics on the official test splits. Avg $M$ denotes the average number of aspects per sentence, together with the proportions of sentences with $M{=}1$ and $M{>}1$.}
    \label{tab:aspect_multiplicity}
    \setlength{\tabcolsep}{2pt}
    \resizebox{\linewidth}{!}{%
    \begin{tabular}{lccccc}
        \toprule
        \textbf{Dataset} & \textbf{Avg $M$} & \textbf{$P(M{=}1)$} & \textbf{$P(M{>}1)$} & \textbf{$P(M{=}2)$} & \textbf{$P(M{>}2)$} \\
        \midrule
        Lap14  & 1.55 & 63.0\% & 37.0\% & 24.8\% & 12.2\% \\
        Rest14 & 1.87 & 47.5\% & 52.5\% & 32.0\% & 20.5\% \\
        Rest15 & 1.49 & 63.6\% & 36.4\% & 26.4\% & 10.0\% \\
        Rest16 & 1.55 & 64.7\% & 35.3\% & 25.5\% & 9.8\% \\
        \bottomrule
    \end{tabular}%
    }
    \vspace{-5mm}
\end{table}

\paragraph{Baselines.}
Our implementation fine-tunes DeBERTa-v3-base ~\cite{he-etal-2023-debertav3}. We compare against three baseline families: (i) \textbf{Structure-aware methods} utilizing external graphs (KGAN-BERT ~\cite{zhong-etal-2023-knowledge}, ASHGAT ~\cite{ouyang-etal-2024-aspect}, DSSK-GAN-BERT ~\cite{liu-etal-2024-graph}, DC-GCN ~\cite{sun-etal-2025-enhancing}, CABiLSTM-BERT ~\cite{he-etal-2025-cabilstm}); (ii) \textbf{Fine-tuning methods} (PConvRoBERTa ~\cite{feng-etal-2023-aspect}, ITGCN ~\cite{shi-etal-2024-aspect}, DeBERTa+RCL ~\cite{jian-etal-2024-retrieval}, Flan-T5-base+Syn-Chain ~\cite{fan-etal-2025-aspect}); and (iii) \textbf{LLMs-based} (Llama-3, Qwen3, GPT-3.5) which we evaluated in a 5-shot in-context learning setup. Full hyperparameter details and baseline descriptions are available in Appendix~\ref{app:implementation_details}.

\begin{table*}[!t]
	\small
	\centering
	\caption{Main results on four ATSA benchmarks (Acc/MF1). DABS reports \textit{mean$\pm$std} over 3 seeds. LLM results are from our 5-shot evaluation setup. Other baseline numbers are taken from prior work. Best within each block is in \textbf{bold} and ``--'' denotes missing values.}
	\label{tab:main_results}
	\setlength{\tabcolsep}{3pt}
	\begin{tabular}{lcc|cc|cc|cc}
		\toprule
		\multirow{2}{*}{\textbf{Models}} &
		\multicolumn{2}{c}{\textbf{Lap14}} &
		\multicolumn{2}{c}{\textbf{Rest14}} &
		\multicolumn{2}{c}{\textbf{Rest15}} &
		\multicolumn{2}{c}{\textbf{Rest16}} \\
		& \textbf{Acc} & \textbf{MF1} & \textbf{Acc} & \textbf{MF1} & \textbf{Acc} & \textbf{MF1} & \textbf{Acc} & \textbf{MF1} \\
		\midrule
		\multicolumn{9}{l}{\textit{Structure-aware methods}}\\
		KGAN-BERT & 82.66 & 78.98 & 87.15 & 82.05 & 86.21 & 74.20 & 92.34 & 81.31 \\
		ASHGAT & 77.91 & 73.89 & 83.45 & 76.55 & 81.06 & 68.72 & 88.96 & 74.02 \\
		DSSK-GAN-BERT & 82.94 & 78.99 & 87.32 & 82.35 & 87.32 & 74.53 & 93.44 & 80.03 \\
		DC-GCN & 80.70 & 77.92 & 87.29 & 81.48 & 84.90 & 72.11 & 92.42 & 79.45 \\
		CABiLSTM-BERT & 77.91 & 73.04 & 83.75 & 75.87 & 82.84 & 66.10 & 86.61 & 73.54 \\
		\midrule
		\multicolumn{9}{l}{\textit{Fine-tuning methods}}\\
		PConvRoBERTa & 83.54 & 80.89 & 89.29 & 84.27 & -- & -- & -- & -- \\
		ITGCN & 82.76 & 79.37 & 88.21 & 82.35 & 87.64 & 75.53 & 93.51 & 79.75 \\
		DeBERTa+RCL & 82.76 & 80.28 & 89.38 & 84.68 & -- & -- & -- & -- \\
		Flan-T5-base+Syn-Chain & 83.22 & 80.04 & 88.39 & 82.79 & 87.82 & \textbf{76.86} & 93.50 & 79.25 \\
		\midrule
		\multicolumn{9}{l}{\textit{Large Language Models (LLM-based, 5-shot)}}\\
		Llama-3-8B-Inst & 64.80 & 55.79 & 82.87 & 71.82 & 84.07 & 75.30 & 79.00 & 68.41 \\
		Qwen3-8B & 68.02 & 62.24 & 82.75 & 73.52 & 81.38 & 70.63 & 86.60 & 73.85 \\
		GPT-3.5 Turbo & 65.48 & 59.74 & 78.85 & 66.50 & 77.63 & 67.28 & 81.30 & 72.91 \\
		\addlinespace
		\hline
		\textbf{DABS (Ours)} & \textbf{84.41}\textsubscript{$\pm$0.15}  & \textbf{81.56}\textsubscript{$\pm$0.29} & \textbf{89.76}\textsubscript{$\pm$0.19} & \textbf{84.87}\textsubscript{$\pm$0.47} & \textbf{89.18}\textsubscript{$\pm$1.20} & 74.06\textsubscript{$\pm$1.61} & \textbf{94.87}\textsubscript{$\pm$0.25} & \textbf{84.38}\textsubscript{$\pm$1.65} \\
		\bottomrule
	\end{tabular}
    \vspace{-2mm}
\end{table*}

\subsection{Quantitative Results}
\label{subsec:quantitative_results}

Table~\ref{tab:main_results} reports the main results. DABS achieves competitive or better Acc/MF1 across all four datasets, with clear gains on Lap14 and Rest16. Compared to structure-aware baselines, DABS remains competitive without requiring external parsers, and it also outperforms strong fine-tuning baselines in several settings. While LLMs show general reasoning ability, 5-shot prompting is less effective for span-level polarity attribution whereby DABS exceeds the best 5-shot LLM results by large margins (\textit{e.g.}, $>10$ pp on Rest16). Additional breakdowns are provided in Appendix~\ref{app:experiment_reproducibility}.

To verify that these gains are robust, we conduct paired statistical tests against an encoder-only baseline. Table~\ref{tab:m7_significance} shows that DABS yields consistent significant improvements. For instance, $\Delta$ MF1 = 6.53 pp on Lap14 and 8.62 pp on Rest16, with $p{<}0.05$ in both cases. Low standard deviations confirm that these gains are consistently driven by our query-readout formulation. Per-seed results are provided in Appendix~\ref{app:significance}.

\begin{table}[t]
    \small
    \centering
    \caption{Matched-backbone comparison on RoBERTa-base. Both models use the same encoder and training protocol. Acc/MF1 (\%) and $\Delta$MF1 relative to the matched encoder-only baseline are reported.}
    \label{tab:matched_backbone_roberta}
    \setlength{\tabcolsep}{5pt}
    \begin{tabular}{lccc}
        \toprule
        \textbf{Dataset} & \textbf{Encoder-only} & \textbf{DABS} & \boldmath$\Delta$\textbf{MF1} \\
        \midrule
        Lap14  & 80.00 / 76.85 & 83.62 / 80.84 & +3.99 \\
        Rest14 & 81.83 / 72.70 & 87.29 / 81.09 & +8.39 \\
        Rest15 & 83.03 / 67.73 & 88.01 / 71.37 & +3.64 \\
        Rest16 & 90.51 / 73.79 & 94.11 / 82.26 & +8.47 \\
        \bottomrule
    \end{tabular}
    \vspace{-3mm}
\end{table}

\paragraph{Matched-backbone validation.}
To isolate framework gains from backbone differences, we additionally evaluate DABS under a matched-backbone setting using RoBERTa-base for both the encoder-only baseline and the full model. Table~\ref{tab:matched_backbone_roberta} reports the results. DABS remains consistently stronger across all four datasets, improving MF1 by +3.64 to +8.47 points over the matched encoder-only baseline. These results indicate that the gains are not attributable to backbone choice, but arise from the proposed single-pass depth substrate and aspect-conditioned readout. Full cross-backbone comparisons are provided in Appendix~\ref{app:backbone_agnostic}.

\begin{table}[t]
	\small
	\centering
	\caption{Paired significance test against the encoder-only baseline (3 matched seeds). $\star$ indicates $p<0.05$.}
	\label{tab:m7_significance}
    \setlength{\tabcolsep}{5pt}
	\begin{tabular}{lccc}
		\toprule
		\textbf{Metric} & \textbf{Lap14} & \textbf{Rest14} & \textbf{Rest16} \\
		\midrule
		MF1\textsubscript{Full} & 81.56\textsubscript{$\pm$0.29} & 84.87\textsubscript{$\pm$0.47} & 84.38\textsubscript{$\pm$1.65} \\
		MF1\textsubscript{Enc}  & 75.03\textsubscript{$\pm$0.43} & 76.33\textsubscript{$\pm$1.24} & 75.76\textsubscript{$\pm$2.49} \\
		$\Delta$ MF1 & 6.53 & 8.54 & 8.62 \\
		$t$-statistic & 17.38 & 16.19 & 17.69 \\
		$p$-value & 0.0033 & 0.0038 & 0.0032 \\
		Significance & $\star$ & $\star$ & $\star$ \\
		\bottomrule
	\end{tabular}
\end{table}

\subsection{Multilingual Generalization}
To assess multilingual generalization beyond English, we evaluate DABS on three non-English SemEval-2016 ABSA benchmarks (French, Russian, and Spanish), all converted into the same ATSA format (Table~\ref{tab:multilingual}). DABS consistently improves over the encoder-only baseline on all three languages in Acc and MF1. Notably, most of the gains come from ACBS-only, suggesting that ACBS transfers well across languages. Instead, DORA primarily serves as an enabling component for \emph{single-pass representation reuse} and for exposing an ordered depth substrate that ACBS can query. Combining both components yields the best overall results, supporting that DORA and ACBS are complementary rather than individually sufficient. As a whole, the ability to navigate the \textbf{depth substrate} is language-agnostic. Detailed per-seed multilingual results and depth-control analyses via region/layer masking are provided in Appendix~\ref{app:multilingual_details}.

\begin{figure*}[t]
	\centering
    \includegraphics[width=0.95\linewidth,keepaspectratio]{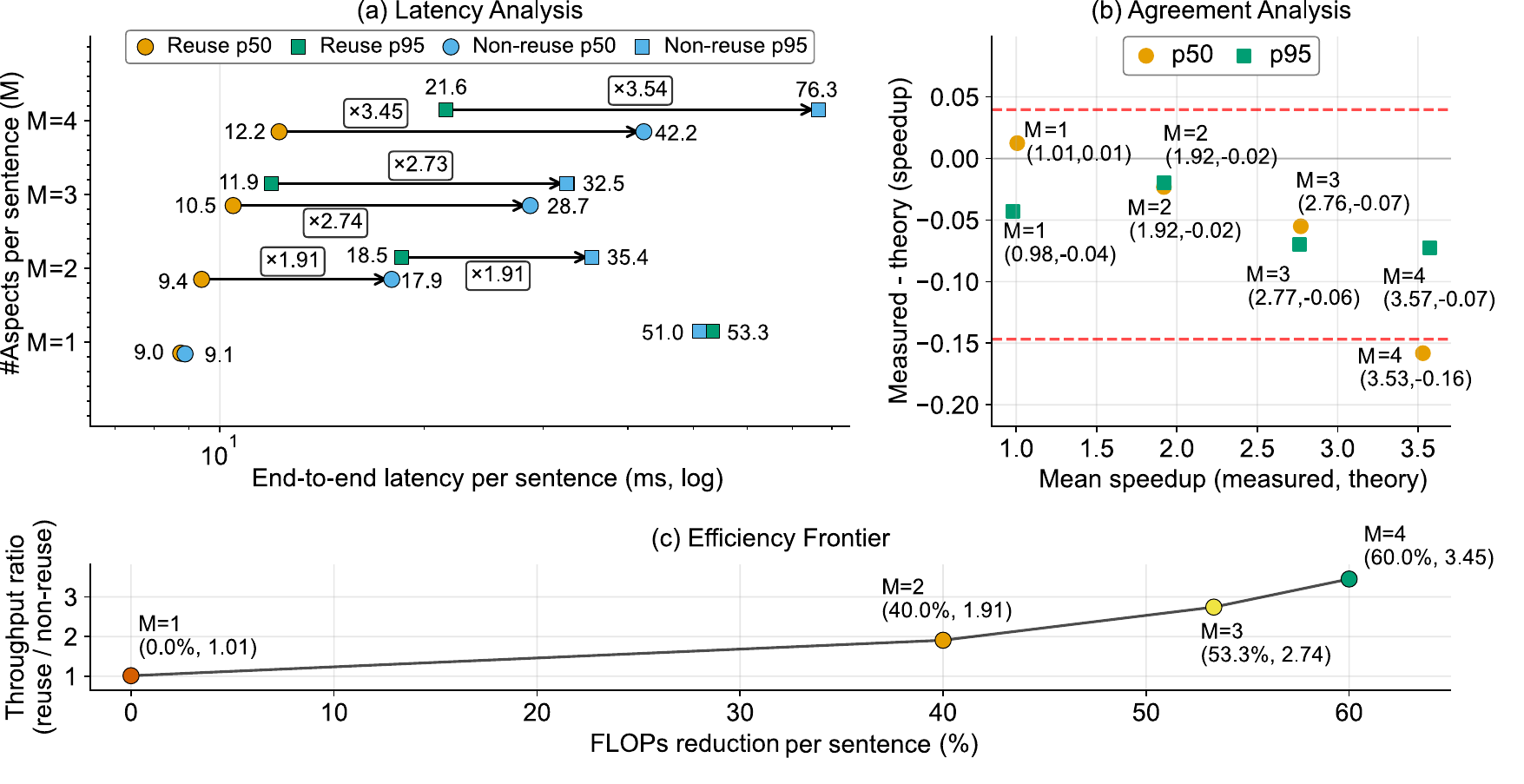}
     \vspace{-3mm}
    \caption{Single-pass reuse under multi-aspect inference ($M$ aspects per sentence), measured with DeBERTa-v3-base on an NVIDIA RTX 5090 GPU. (a) Reuse reduces p50/p95 latency compared to non-reuse, with larger gains as $M$ increases. (b) Measured speedups match the cost-model predictions; red dashed lines mark the 95\% agreement bounds. (c) FLOPs reduction leads to higher throughput, forming a clear efficiency frontier as $M$ grows.}
	\label{fig:multi_aspect}
    \vspace{-3mm}
\end{figure*}

\subsection{Efficiency and Reuse Analysis}
\label{subsec:efficiency}

A central advantage of DABS lies in its ability to \textbf{amortize representational cost} across multiple aspect queries while allowing each aspect to independently allocate depth based on its semantic demands. In DABS, the expensive encoder pass and reusable depth substrate are computed once per sentence, and additional aspects incur only the marginal cost of lightweight depth allocation and evidence selection. In contrast, non-reuse baselines re-encode the sentence for each aspect, repeatedly paying the full cost of deep representation regardless of aspect complexity. To isolate the fixed construction overhead, Table~\ref{tab:sa2_efficiency_extended} in Appendix reports component-wise metrics under a single-aspect setting ($M{=}1$). As $M$ increases, this fixed cost is rapidly amortized, leading to the significant latency and throughput gains illustrated in Figure~\ref{fig:multi_aspect}.

\paragraph{End-to-end latency.}
Beyond the reuse/non-reuse comparison, we evaluate end-to-end latency against KGAN-BERT, a representative structure-aware baseline with a complete runnable pipeline. Table~\ref{tab:e2e_latency_kgan} shows that DABS consistently reduces p95/p99 latency under offered load, indicating that the benefit of single-pass reuse is not limited to the internal amortization analysis in Figure~\ref{fig:multi_aspect}, but also yields end-to-end gains against a competitive baseline.

\begin{table}[t]
    \small
    \centering
    \caption{End-to-end tail latency (ms) under offered load, comparing DABS with KGAN-BERT. Lower is better.}
    \label{tab:e2e_latency_kgan}
    \setlength{\tabcolsep}{3pt}
    \resizebox{\linewidth}{!}{%
    \begin{tabular}{lcccccc}
        \toprule
        \multirow{2}{*}{\makecell{\textbf{Offered}\\\textbf{QPS}}} &
        \multicolumn{2}{c}{\textbf{KGAN-BERT}} &
        \multicolumn{2}{c}{\textbf{DABS (Ours)}} &
        \multirow{2}{*}{\makecell{\textbf{p95}\\\textbf{Speedup}}} &
        \multirow{2}{*}{\makecell{\textbf{p99}\\\textbf{Speedup}}} \\
        \cmidrule(lr){2-3} \cmidrule(lr){4-5}
        & \textbf{p95} & \textbf{p99} & \textbf{p95} & \textbf{p99} & \\
        \midrule
        60  & 269.48 & 328.95 & 131.58 & 176.21 & 2.05$\times$ & 1.87$\times$ \\
        100 & 286.43 & 345.73 & 188.10 & 203.73 & 1.52$\times$ & 1.70$\times$ \\
        \bottomrule
    \end{tabular}
    }
\end{table}

\begin{table}[t]
	\small
	\centering
	\caption{Multilingual transfer on SemEval-2016 ABSA benchmarks (French/Russian/Spanish). Results are \textit{mean$\pm$std} over 3 seeds in Acc. and MF1.}
	\label{tab:multilingual}
	\setlength{\tabcolsep}{1pt}
	\resizebox{\linewidth}{!}{%
	\begin{tabular}{lcccccc}
		\toprule
		\multirow{2}{*}{\textbf{Methods}} & \multicolumn{2}{c}{\textbf{French}} & \multicolumn{2}{c}{\textbf{Russian}} & \multicolumn{2}{c}{\textbf{Spanish}} \\
		\cmidrule(lr){2-3} \cmidrule(lr){4-5} \cmidrule(lr){6-7}
		& Acc & MF1 & Acc & MF1 & Acc & MF1 \\
		\midrule
		Baseline  & 85.64\textsubscript{$\pm$0.62} & 74.67\textsubscript{$\pm$0.63} & 84.50\textsubscript{$\pm$0.89} & 73.29\textsubscript{$\pm$0.68} & 88.95\textsubscript{$\pm$0.57} & 71.73\textsubscript{$\pm$1.94} \\
		DORA-only & 85.03\textsubscript{$\pm$0.32} & 73.97\textsubscript{$\pm$0.16} & 84.88\textsubscript{$\pm$0.37} & 72.87\textsubscript{$\pm$0.30} & 88.29\textsubscript{$\pm$0.86} & 71.25\textsubscript{$\pm$2.52} \\
		ACBS-only & 88.56\textsubscript{$\pm$0.87} & 79.17\textsubscript{$\pm$1.68} & 88.62\textsubscript{$\pm$0.64} & 77.75\textsubscript{$\pm$1.83} & 92.71\textsubscript{$\pm$0.67} & 77.97\textsubscript{$\pm$1.10} \\
		\midrule
		\textbf{DABS} & \textbf{89.28}\textsubscript{$\pm$0.89} & \textbf{80.94}\textsubscript{$\pm$1.34} & \textbf{89.39}\textsubscript{$\pm$0.32} & \textbf{79.89}\textsubscript{$\pm$0.14} & \textbf{93.14}\textsubscript{$\pm$0.86} & \textbf{80.05}\textsubscript{$\pm$1.44} \\
		\bottomrule
	\end{tabular}
	}
     \vspace{-3mm}
\end{table}

Figure~\ref{fig:multi_aspect}(a) shows this effect in terms of latency as the number of aspects $M$ increases. While non-reuse models exhibit near-linear growth due to repeated deep encoding, DABS shows substantially flatter latency curves. At $M{=}4$, DABS substantially reduces both median and tail latency over the non-reuse baseline. This shows that depth-related computation is largely paid once, and subsequent aspects primarily incur the cost of selective depth querying. Figure~\ref{fig:multi_aspect}(b) compares the observed speedup against the theoretical amortization predicted by the cost model. We observe close agreement between measured and cost-model speedup. Deviations remain small for most settings but increase at higher $M$, with the largest gap in $p_{50}$ at $M{=}4$. Tail-latency ($p_{95}$) deviations also increase from $M\ge3$ but remain comparatively stable.

Finally, Figure~\ref{fig:multi_aspect}(c) summarizes the resulting cost-throughput tradeoff. As $M$ increases, the proportion of computation attributed to shared depth construction dominates, leading to increasing FLOPs reduction (up to 60.0\% at $M{=}4$) and higher throughput ratios (up to $3.45\times$). Together, these results show that DABS separates fixed deep encoding from aspect-specific depth selection, enabling efficient multi-aspect inference without sacrificing fine-grained accuracy.

\begin{figure}[t]
    \centering
    \includegraphics[width=0.9\linewidth,keepaspectratio]{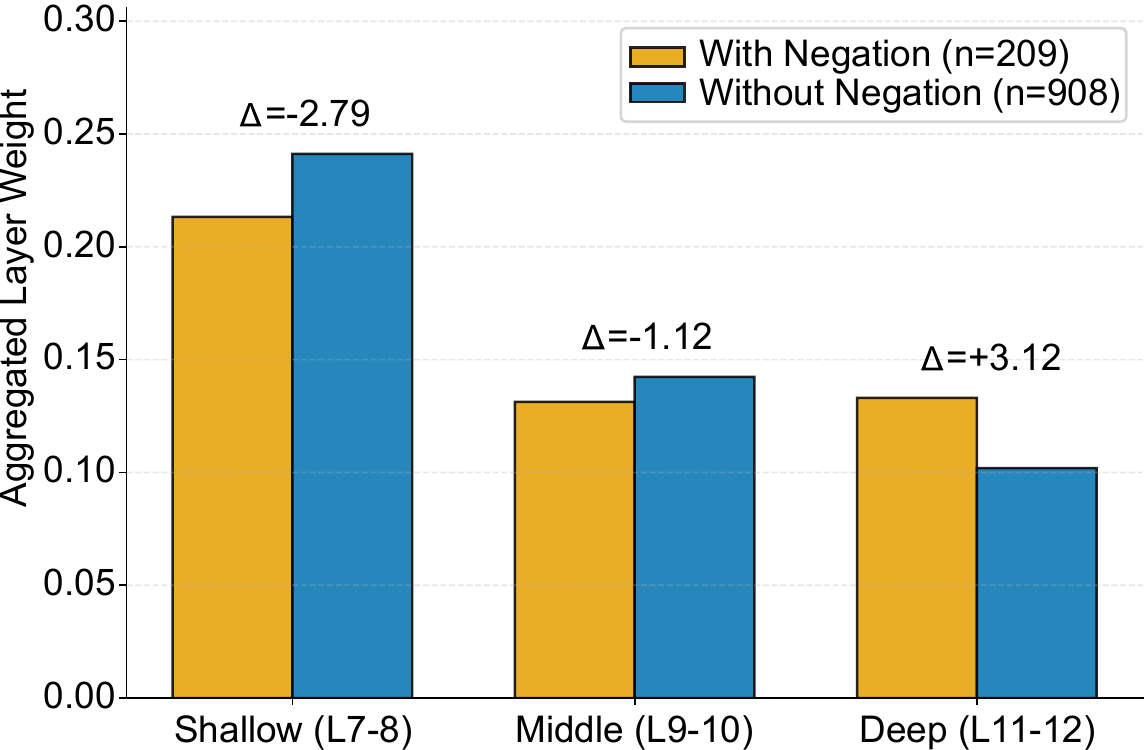}
    \caption{Depth allocation under negation. Aggregated depth-region weights for sentences with vs.\ without negation, showing a shift toward deeper layers.}
    \label{fig:negation_depth}
\end{figure}
\begin{figure*}[t]
	\centering
    \includegraphics[width=0.96\textwidth,keepaspectratio]{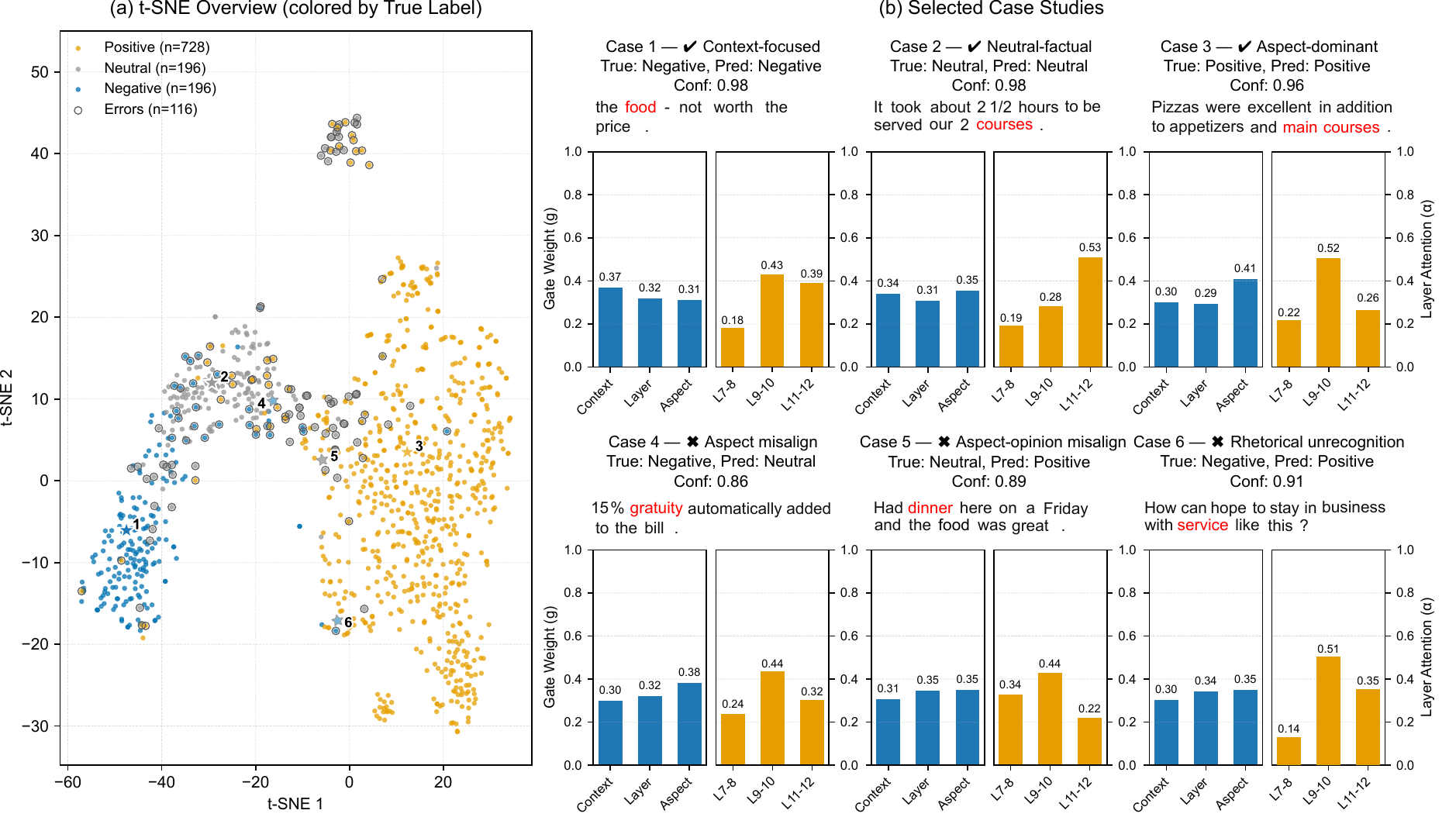}
	\caption{Learned representations and selection behavior. (a) t-SNE of instance embeddings by gold label. (b) Case studies showing fusion-gate weights and depth level $L$.}	
	\label{fig:tsne_cases}
    \vspace{-5mm}
\end{figure*}

\paragraph{Why querying the last $K$ layers.} 
A setting of $K{=}6$ balances expressiveness and overhead. The upper Transformer layers typically encode higher-level compositional and discourse interactions, while still retaining sufficient lexical grounding for span-level polarity. Restricting the substrate to the last $K$ layers also makes depth controls (region/layer masking) easier to interpret and keeps the additional DORA parameters and memory footprint bounded. Table~\ref{tab:k_sensitivity_main} also shows that larger budgets bring smaller gains, while p50 latency increases across the four benchmarks. Together, these results support $K{=}6$ as a stable choice.

\begin{table}[t]
    \small
    \centering
    \caption{Sensitivity to the depth budget. Values are reported as Acc, MF1, and p50 latency.}
    \label{tab:k_sensitivity_main}
    \setlength{\tabcolsep}{3pt}
    \resizebox{\linewidth}{!}{%
    \begin{tabular}{llcccc}
        \toprule
        \textbf{Dataset} & \textbf{Metric} & \textbf{K=2} & \textbf{K=4} & \textbf{K=6} & \textbf{K=8} \\
        \midrule
        \multirow{3}{*}{Lap14}
          & Acc & 83.20$_{\pm0.92}$ & 83.36$_{\pm0.73}$ & \textbf{84.41$_{\pm0.15}$} & 83.88$_{\pm0.24}$ \\
          & MF1 & 80.27$_{\pm1.02}$ & 80.55$_{\pm1.18}$ & \textbf{81.56$_{\pm0.29}$} & 80.78$_{\pm0.17}$ \\
          & p50 & \textbf{11.64$_{\pm0.37}$} & 12.11$_{\pm0.28}$ & 12.51$_{\pm0.17}$ & 12.95$_{\pm0.86}$ \\
        \midrule
        \multirow{3}{*}{Rest14}
          & Acc & 89.38$_{\pm0.31}$ & 89.38$_{\pm0.27}$ & \textbf{89.76$_{\pm0.19}$} & 89.47$_{\pm0.70}$ \\
          & MF1 & 84.05$_{\pm0.63}$ & 83.89$_{\pm0.80}$ & \textbf{84.87$_{\pm0.47}$} & 84.47$_{\pm1.40}$ \\
          & p50 & \textbf{12.18$_{\pm0.43}$} & 12.42$_{\pm0.36}$ & 12.70$_{\pm0.55}$ & 12.92$_{\pm0.67}$ \\
        \midrule
        \multirow{3}{*}{Rest15}
          & Acc & 88.44$_{\pm0.70}$ & 88.68$_{\pm1.23}$ & 89.18$_{\pm1.20}$ & \textbf{89.28$_{\pm0.75}$} \\
          & MF1 & 73.08$_{\pm1.35}$ & 73.68$_{\pm2.08}$ & 74.06$_{\pm1.61}$ & \textbf{74.61$_{\pm1.48}$} \\
          & p50 & \textbf{12.43$_{\pm0.52}$} & 12.79$_{\pm0.41}$ & 13.03$_{\pm0.61}$ & 13.31$_{\pm0.58}$ \\
        \midrule
        \multirow{3}{*}{Rest16}
          & Acc & 93.78$_{\pm1.13}$ & 94.05$_{\pm0.47}$ & \textbf{94.87$_{\pm0.25}$} & 93.84$_{\pm0.50}$ \\
          & MF1 & 82.56$_{\pm2.72}$ & 83.53$_{\pm1.91}$ & \textbf{84.38$_{\pm1.65}$} & 82.33$_{\pm2.61}$ \\
          & p50 & \textbf{12.12$_{\pm0.24}$} & 12.46$_{\pm0.47}$ & 12.78$_{\pm0.34}$ & 13.09$_{\pm0.65}$ \\
        \bottomrule
    \end{tabular}
    }
    \vspace{-5mm}
\end{table}

\subsection{Qualitative Analysis}
\label{subsec:qual}

To verify if DABS learns meaningful linguistic structures, we analyze how depth selection shifts in the presence of negation.

\textbf{Negation induces deeper reading.} 
Negation is a canonical compositional phenomenon where correct sentiment depends on combining multiple cues (\textit{e.g.}, ``\textit{not}'' reversing ``\textit{good}'') rather than single lexical triggers.
As shown in Figure~\ref{fig:negation_depth}, which aggregates data from 1,117 test samples ($n=209$ negated, $n=908$ non-negated), the learned depth selector exhibits a distinct shift.
Specifically, the model reduces weight on surface-level features ($\Delta=-2.79$ pp in Shallow layers L\textsubscript{7-8}) and actively recruits deeper abstractions ($\Delta=+3.12$ pp in Deep layers L\textsubscript{11-12}).
This pattern provides direct evidence that the selector does not mix layers arbitrarily. Instead, it functions as a semantic cursor, automatically shifting from \textbf{surface-level lexical matching} to \textbf{deep compositional logic} only when the linguistic complexity requires it.

\textbf{Case Studies.}
Figure~\ref{fig:tsne_cases}(a) shows that the learned instance embeddings form label-aligned regions with remaining overlap concentrated around neutral and hard examples. Figure~\ref{fig:tsne_cases}(b) shows how fusion gates and depth allocation correlate with both correct predictions and failures. For successful cases, Case~1 (negation) is context-focused, while Case~3 (explicit sentiment) becomes aspect-dominant and concentrates on middle layers (L\textsubscript{9-10}). The neutral-factual Case~2 shifts $\bm{\alpha}$ toward deeper layers (L\textsubscript{11-12}), consistent with higher contextual demand. Notably, high-confidence errors (Cases~4-6) expose systematic limits in which the model could mis-handle \emph{implicit/pragmatic negativity} (Case~4), confuse \emph{target-opinion alignment} when multiple nearby candidates exist (Case~5), and fail on \emph{rhetorical polarity} without explicit sentiment cues (Case~6). Overall, the gates and $\bm{\alpha}$ diagnostically align with compositional needs for explicit sentiment, while errors track implicit or misaligned evidence.

%gates and $\bm{\alpha}$ are a diagnostic signal by aligning with compositional needs when sentiment is explicit, while errors concentrate on implicit or misaligned evidence.

\subsection{Ablation Study}
\label{subsec:ablation}
To validate the contributions of DORA and ACBS, we perform an ablation on $\mathrm{MF1}$ (Table~\ref{tab:ablation_mf1}). Accuracy ablations follow similar trends and are reported in Appendix~\ref{app:ablation_acc_app} (Table \ref{tab:ablation_acc}).

\begin{table}[t]
	\small
	\centering
	\caption{Ablations on $\mathrm{MF1}_{\text{mean}}$. Values are reported as MF1$^{\Delta}$, where $\Delta$ is the change (pp) from the full model.}
	\label{tab:ablation_mf1}
	\setlength{\tabcolsep}{2pt}
	\resizebox{\linewidth}{!}{%
		\begin{tabular}{l|cccc|c}
			\toprule
			\multirow{2}{*}{Config} & Lap14 & Rest14 & Rest15 & Rest16 & Avg \\
			& MF1$^{\Delta}$ & MF1$^{\Delta}$ & MF1$^{\Delta}$ & MF1$^{\Delta}$ & $\Delta$ \\
			\midrule
			\textbf{DABS (Full)} & \textbf{81.56}$^{\scriptsize 0.00}$ & \textbf{84.87}$^{\scriptsize 0.00}$ & \textbf{74.06}$^{\scriptsize 0.00}$ & \textbf{84.38}$^{\scriptsize 0.00}$ & 0.00 \\
			\midrule
			\textit{w/o Regularizers} & & & & & \\
			~- Sparsity          & 79.84$^{\scriptsize \reddown1.72}$ & 83.68$^{\scriptsize \reddown1.19}$ & 72.68$^{\scriptsize \reddown1.38}$ & 82.25$^{\scriptsize \reddown2.13}$ & $\reddown$1.61 \\
			~- Span Masking  & 79.77$^{\scriptsize \reddown1.79}$ & 82.77$^{\scriptsize \reddown2.10}$ & 72.62$^{\scriptsize \reddown1.44}$ & 84.37$^{\scriptsize \reddown0.01}$ & $\reddown$1.34 \\
			~- Gate Entropy      & 79.68$^{\scriptsize \reddown1.88}$ & 84.14$^{\scriptsize \reddown0.73}$ & 73.08$^{\scriptsize \reddown0.98}$ & 83.33$^{\scriptsize \reddown1.05}$ & $\reddown$1.16 \\
			\midrule
			\textit{w/o DORA} & & & & & \\
			~- DepthGRU        & 80.81$^{\scriptsize \reddown0.75}$ & 83.45$^{\scriptsize \reddown1.42}$ & 74.46$^{\scriptsize \greenup0.40}$ & 80.73$^{\scriptsize \reddown3.65}$ & $\reddown$1.36 \\
			~- LCP (Pooling)   & 79.31$^{\scriptsize \reddown2.25}$ & 82.74$^{\scriptsize \reddown2.13}$ & 75.29$^{\scriptsize \greenup1.23}$ & 80.37$^{\scriptsize \reddown4.01}$ & $\reddown$1.79 \\
			\midrule
			\textit{w/o ACBS} & & & & & \\
			~- Token Sel.   & 78.25$^{\scriptsize \reddown3.31}$ & 83.14$^{\scriptsize \reddown1.73}$ & 71.84$^{\scriptsize \reddown2.22}$ & 80.57$^{\scriptsize \reddown3.81}$ & $\reddown$2.77 \\
			~- Layer Sel.   & 79.95$^{\scriptsize \reddown1.61}$ & 82.35$^{\scriptsize \reddown2.52}$ & 72.63$^{\scriptsize \reddown1.43}$ & 83.04$^{\scriptsize \reddown1.34}$ & $\reddown$1.73 \\
			~- Gated Fusion & 79.06$^{\scriptsize \reddown2.50}$ & 81.68$^{\scriptsize \reddown3.19}$ & 73.29$^{\scriptsize \reddown0.77}$ & 82.02$^{\scriptsize \reddown2.36}$ & $\reddown$2.21 \\
			\bottomrule
		\end{tabular}%
	}
    \vspace{-5mm}
\end{table}

\textbf{Impact of Regularizers.} Removing sparsity causes the largest drop on Rest16 (-2.13 pp), and on average -1.61 confirming that filtering redundant context is crucial for longer reviews. Gate entropy is essential for preventing mode collapse in the fusion layer as removing it degrades performance consistently across all datasets.

\textbf{Impact of DORA.} The removal of DepthGRU results in a 1.36 pp drop on average, validating that multi-level aggregation is necessary to capture semantic abstractions in complex sentences. LCP is similarly critical, without it, the model fails to capture local compositional cues (\textit{\textit{e.g.}}, negations).

\textbf{Impact of ACBS.} This is the most critical module. Removing token selection yields the largest average degradation (-2.77 pp), proving that independent token gating is essential for aggregating distributed sentiment evidence. Layer selection is equally vital, as different aspects require different levels of semantic abstraction (see Section~\ref{subsec:qual}).

\textbf{Impact of DepthGRU.} This ablation isolates DepthGRU within DORA by removing only the cross-layer recurrence while keeping all other components fixed. Table~\ref{tab:depth_gru_ablation_k6a} shows that removing DepthGRU decreases Macro-F1 on all stress-test splits. Most notably on Multi-Aspect Conflict (MA-C) and Complex Negation, while Accuracy can remain relatively stable, suggesting DepthGRU primarily improves hard/minority-case discrimination. See Appendix \ref{app:cross_dataset_controls} for completeness.

\begin{table}[t]
	\centering
	\caption{DepthGRU ablation on cross-dataset stress-test splits ($K{=}6$). Results are over 3 seeds (Acc/MF1).}	
	\label{tab:depth_gru_ablation_k6a}
	\setlength{\tabcolsep}{3pt}
	\resizebox{\linewidth}{!}{%
		\begin{tabular}{llcc}
			\toprule
			\textbf{Dataset} & \textbf{Setting} & \textbf{Acc} & \textbf{MF1} \\
			\midrule
			\multirow{2}{*}{\parbox{2.1cm}{Long \\ Sentences}}
			& w DepthGRU & 82.95\textsubscript{$\pm$ 0.91} & 78.44\textsubscript{$\pm$ 0.76} \\
			& w/o DepthGRU & 82.61\textsubscript{$\pm$ 1.57} & 76.93\textsubscript{$\pm$ 2.14} \\
			\midrule
			\multirow{2}{*}{\parbox{2.1cm}{Multi-Aspect \\ Conflict}}
			& w DepthGRU & 95.40\textsubscript{$\pm$ 1.15} & 76.88\textsubscript{$\pm$ 10.80} \\
			& w/o DepthGRU & 94.83\textsubscript{$\pm$ 0.57} & 71.32\textsubscript{$\pm$  6.29} \\
			\midrule
			\multirow{2}{*}{\parbox{2.1cm}{Complex \\ Negation}}
			& w DepthGRU  & 83.81\textsubscript{$\pm$ 2.18} & 80.62\textsubscript{$\pm$ 1.49} \\
			& w/o DepthGRU & 82.86\textsubscript{$\pm$ 2.86} & 77.69\textsubscript{$\pm$ 3.36} \\
			\bottomrule
	\end{tabular}}
    \vspace{-5mm}
\end{table}

%%%%%%%%%%%%%%%%%%%%%%%%%%%%%%%%%%%%

\section{Conclusion}
This paper addresses the efficiency-expressiveness tradeoff in ATSA by framing inference as a resource allocation problem. We propose DABS, a single-pass framework that reuses a shared encoder pass while enabling lightweight, aspect-conditioned readout. Across four benchmarks, DABS achieves competitive Accuracy and Macro-F1 while reducing end-to-end computation by up to 60\% in multi-aspect settings. These results show that explicitly querying Transformer depth, rather than treating it as a static feature pool, enables efficient reuse and targeted sentiment reasoning.
\clearpage

\section*{Acknowledgment}
This work is financially supported by VinUniversity under Grant No. VUNI.2526.AREP.004.

\section*{Limitations}
While DABS shows that \textit{single-pass reuse} and \textit{budget-aware depth querying} can improve the efficiency-expressiveness tradeoff in multi-aspect ATSA, its current formulation is constrained by three practical constraints. Specifically, \textbf{(i) Amortization-dependent gains.} Since DABS improves efficiency by amortizing a shared encoder pass across aspects, rather than by reducing backbone computation itself, its advantage is most pronounced in true multi-aspect settings and naturally smaller when $M{=}1$; \textbf{(ii) Retention cost of the depth substrate.} Keeping the last $K$ layers as a reusable substrate incurs a fixed compute and memory cost that increases with sequence length, model size, and $K$, and finally \textbf{(iii) Task scope.} We evaluate DABS only in span-given ATSA with gold aspect spans. Extending the framework to settings with latent, implicit, or jointly extracted aspects is left to future work.

\bibliography{references}

\clearpage

\appendix

\section{Dataset Statistics}
\label{app:dataset_stats}

Section~\ref{subsec:setup} describes the datasets and evaluation protocol. For completeness, Table~\ref{tab:dataset_statistics_augmented} reports the class distributions of the official train/test splits. 
We include the laptop and restaurant benchmarks from the SemEval-2014 Task 4 ~\cite{pontiki-etal-2014-semeval}, SemEval-2015 Task 12 ~\cite{pontiki-etal-2015-semeval}, and SemEval-2016 Task 5 ~\cite{pontiki-etal-2016-semeval} challenges. The multilingual datasets (Spanish, French, and Russian) are also sourced from the SemEval-2016 subtasks ~\cite{pontiki-etal-2016-semeval}. All counts are at the \emph{aspect-instance} level (not sentence level), grouped by sentiment label (\emph{positive}, \emph{neutral}, \emph{negative}).

\begin{table}[!ht]
	\centering
    \small
	\caption{Dataset statistics.}
	\label{tab:dataset_statistics_augmented}
    \setlength{\tabcolsep}{5pt}
    \begin{tabular}{lccccccc}
        \toprule
        \textbf{Dataset} & \multicolumn{2}{c}{\textbf{Positive}} &
        \multicolumn{2}{c}{\textbf{Neutral}} & \multicolumn{2}{c}{\textbf{Negative}} \\
        \cmidrule(lr){2-3}\cmidrule(lr){4-5}\cmidrule(lr){6-7}
        & \textbf{Train} & \textbf{Test} & \textbf{Train} & \textbf{Test} &
        \textbf{Train} & \textbf{Test} \\
        \midrule
        Lap14   & 994  & 341 & 464 & 169 & 870 & 128 \\
        Rest14  & 2164 & 728 & 637 & 196 & 807 & 196 \\
        Rest15  & 963  & 353 & 36  & 37  & 280 & 207 \\
        Rest16  & 1319 & 483 & 72  & 32  & 489 & 135 \\
        \midrule
        Spanish  & 1368 & 521 & 89  & 34  & 479 & 176 \\
        French  & 901  & 364 & 116 & 69  & 753 & 285 \\
        Russian  & 2453 & 670 & 224 & 92  & 481 & 207 \\
        \bottomrule
    \end{tabular}%
\end{table}

\paragraph{Label imbalance and metric sensitivity.}
Table~\ref{tab:dataset_statistics_augmented} highlights a non-trivial label imbalance that is particularly pronounced in the restaurant benchmarks, where the \emph{neutral} class can be extremely small (\textit{e.g.}, Rest15/Rest16).
This imbalance makes macro-F1 more sensitive to minority-class recall and can increase variance across random seeds, especially when evaluation subsets further reduce neutral counts. Accordingly, we report both Accuracy (Acc) and Macro-F1 (MF1), as MF1 is more sensitive to minority-class performance.
Since the SemEval benchmarks provide only official train/test splits without a standard development set, we train on the official training split and evaluate on the official test split. 
All statistics are reported at the \emph{aspect-instance} level, meaning a single multi-aspect sentence contributes multiple labeled instances. This is aligned with our multi-query formulation where aspects are treated as parallel queries over shared context.

\section{Implementation Details}
\label{app:implementation_details}

This appendix summarizes span alignment, hyperparameters, optimization settings, and baseline protocols used in our experiments. Unless otherwise noted, the backbone encoder remains aspect-agnostic so that a single sentence encoding can be reused across all aspects in the same sentence.

\subsection{Tokenization and Span Alignment}

We tokenized sentences using the DeBERTa-v3-base tokenizer. SemEval aspect annotations are provided as contiguous spans in the raw sentence. We mapped each gold span to subword indices via offset mappings returned by the tokenizer. When an aspect term was split into multiple subwords, we used the full contiguous subword span and treated it as the aspect interval $[i,j]$ (Section~\ref{subsec:problem} in the main paper). Aspect vectors were computed by mean pooling over the enhanced sequence representations $E$ within the aspect span (Section~\ref{subsec:acbs} in the main paper).

\subsection{Experimental Environment}

All experiments were run on an AMD Ryzen 9 9950X CPU and an NVIDIA GeForce RTX 5090 GPU with $32\,\mathrm{GB}$ memory.
The software stack used PyTorch~2.8.0, CUDA~12.8, and Transformers~4.52.0. We fixed random seeds to $\{42,123,456\}$ unless stated otherwise.

\subsection{Model Hyperparameters}

We used DeBERTa-v3-base as the default backbone encoder. For depth querying, DORA exposed a depth substrate built from the last $K{=}6$ Transformer layers. In DORA, the Local Context Preservation (LCP) module applied three depthwise-separable convolution branches with kernel sizes $k\in\{1,3,5\}$, keeping short-range compositional cues (\textit{e.g.}, negation, degree modifiers) explicit in a reusable substrate.

In ACBS, we applied one multi-head self-attention layer over $E$ to obtain the reusable context $C$. The token selector used a two-layer MLP with shape $2d\!\rightarrow\! d\!\rightarrow\! 1$ (GELU), the depth selector used $2d\!\rightarrow\! d\!\rightarrow\! K$, and the fusion gate used $3d\!\rightarrow\! d\!\rightarrow\! 3$. We set $\varepsilon=10^{-6}$ in Eq.~\eqref{eq:token_select}. Temperatures were fixed to $\tau_\alpha=\tau_g=1.0$ in Eqs.~\eqref{eq:alpha} and \eqref{eq:fusion}. Unless otherwise noted, the maximum sequence length was 256.

\subsection{Optimization and Regularization}
We train on the official training split for up to 30 epochs with a batch size of 32. Because these benchmarks do not provide a standard development set, for each run we report the best test-set score within the training budget.

Optimization used AdamW with a shared learning rate, selected from \{\texttt{2e-5}, \texttt{3e-5}\} depending on the dataset.
We used the default linear schedule in Transformers, with dataset-specific warmup ratios. Mixed-precision training (\texttt{bf16}) was enabled, and gradient clipping was applied with $\|g\|_2 \le 1$.

Dropout was set to 0.1 for the encoder and most DORA/ACBS internal modules, while the final classifier used 0.2. For the objective in Eq.~\eqref{eq:total-loss}, we set $\lambda_s=\lambda_m=10^{-3}$. For the entropy weight $\lambda_{\text{ent}}$, we used a fixed value of $10^{-2}$.

More implementation details and configuration settings are available at \url{https://github.com/panzhzh/acl-dabs}.

\subsection{Baseline Protocols}

We compared against three baseline groups as described in Section~\ref{subsec:setup}(in the main paper):
(i) structure-aware methods that inject syntactic/knowledge structure,
(ii) fine-tuning paradigms that treat ATSA as supervised classification, and
(iii) LLMs evaluated with 5-shot in-context learning.

\subsubsection{LLM 5-shot In-Context Learning}

For each test instance $(s^\ast,a^\ast)$, we retrieved $k{=}5$ demonstrations from the in-domain training split using a convex combination of sentence similarity and aspect-term similarity.
We computed cosine similarities between normalized embeddings.
Let
$u_i=\cos(\phi(s^\ast),\phi(s_i))$ denote sentence similarity and
$v_i=\cos(\phi(a^\ast),\phi(a_i))$ denote aspect similarity.
Candidates were ranked by
\[
\mathrm{score}_i = \lambda\, v_i + (1-\lambda)\, u_i,\quad \lambda=0.6,
\]
and we used the top-$k$ items as demonstrations.

Here, $\phi(\cdot)$ is a frozen embedding function used only for retrieval.
To avoid introducing extra models, we used mean-pooled last-layer embeddings from the same DeBERTa-v3-base encoder (without task-specific heads) as $\phi(\cdot)$. Prompts followed a fixed template and required the model to output exactly one label from \{\texttt{positive}, \texttt{neutral}, \texttt{negative}\}. Decoding used temperature $=0.1$, top-$p=1.0$, and $\mathrm{max\_new\_tokens}=1$. We did not update model parameters for LLM baselines.

The 5-shot prompt template is shown below.

\begin{tcolorbox}[colback=gray!3,colframe=gray!50,title=Aspect-Term Sentiment Prompt (5-shot),left=2mm,right=2mm,boxsep=1mm]
	\begin{small}
		You are an aspect-term sentiment classifier. Your task is to determine the sentiment polarity of a given aspect term within a sentence.\\
		
		Follow the rules below strictly:\\
		1. Output must be exactly one label chosen from: \texttt{positive}, \texttt{neutral}, \texttt{negative}.\\
		2. Use lowercase only. Do not output any additional words, punctuation, or whitespace.\\
		3. Base your decision solely on the provided sentence and the marked aspect term.\\
		4. Do not provide explanations or reformulations.
		
		\medskip
		\textbf{Examples}\\
		\texttt{Sentence:} \{s$_1$\}\quad \texttt{Aspect:} \{a$_1$\}\quad \texttt{Label:} \{y$_1$\}\\
		\dots\\
		\texttt{Sentence:} \{s$_k$\}\quad \texttt{Aspect:} \{a$_k$\}\quad \texttt{Label:} \{y$_k$\}
		
		\medskip
		\textbf{Query}\\
		\texttt{Sentence:} \{s$^\ast$\}\quad \texttt{Aspect:} \{a$^\ast$\}\quad \texttt{Label:}
	\end{small}
\end{tcolorbox}

\section{Extended Experiments and Reproducibility}
\label{app:experiment_reproducibility}

\begin{table}[t]
	\small
	\centering
	\caption{Extended effectiveness (mean$\pm$std over 3 seeds): MF1/Acc and $\Delta$ (pp) relative to the encoder-only baseline. \textbf{Bold} indicates the best result per dataset.}
	\label{tab:sa1_effectiveness_extended}
	\setlength{\tabcolsep}{2pt}
	\resizebox{\linewidth}{!}{%
		\begin{tabular}{llcccc}
			\toprule
			Dataset & Config & MF1 & $\Delta$ MF1 & Acc & $\Delta$ Acc \\
			\midrule
			Lap14 & Encoder-only & 75.03\textsubscript{$\pm$0.43} & 0.00 & 79.16\textsubscript{$\pm$0.24} & 0.00 \\
			Lap14 & DORA-only    & 75.72\textsubscript{$\pm$0.88} & 0.69 & 79.63\textsubscript{$\pm$0.51} & 0.47 \\
			Lap14 & ACBS-only    & 79.97\textsubscript{$\pm$0.73} & 4.94 & 82.62\textsubscript{$\pm$0.55} & 3.46 \\
			Lap14 & Full         & \textbf{81.56\textsubscript{$\pm$0.29}} & 6.53 & \textbf{84.41\textsubscript{$\pm$0.16}} & 5.25 \\
			\midrule
			Rest14 & Encoder-only & 76.33\textsubscript{$\pm$1.24} & 0.00 & 84.48\textsubscript{$\pm$0.49} & 0.00 \\
			Rest14 & DORA-only    & 73.18\textsubscript{$\pm$0.23} & -3.15 & 83.02\textsubscript{$\pm$0.31} & -1.46 \\
			Rest14 & ACBS-only    & 83.15\textsubscript{$\pm$1.68} & 6.82 & 88.81\textsubscript{$\pm$1.01} & 4.33 \\
			Rest14 & Full         & \textbf{84.87\textsubscript{$\pm$0.47}} & 8.54 & \textbf{89.76\textsubscript{$\pm$0.19}} & 5.28 \\
			\midrule
			Rest15 & Encoder-only & 71.39\textsubscript{$\pm$1.12} & 0.00 & 85.85\textsubscript{$\pm$0.91} & 0.00 \\
			Rest15 & DORA-only    & 66.60\textsubscript{$\pm$4.87} & -4.79 & 84.56\textsubscript{$\pm$1.23} & -1.29 \\
			Rest15 & ACBS-only    & 72.60\textsubscript{$\pm$0.39} & 1.21 & 88.19\textsubscript{$\pm$0.85} & 2.34 \\
			Rest15 & Full         & \textbf{74.06\textsubscript{$\pm$1.61}} & 2.67 & \textbf{89.18\textsubscript{$\pm$1.20}} & 3.33 \\
			\midrule
			Rest16 & Encoder-only & 75.76\textsubscript{$\pm$2.49} & 0.00 & 91.00\textsubscript{$\pm$0.43} & 0.00 \\
			Rest16 & DORA-only    & 72.82\textsubscript{$\pm$1.50} & -2.94 & 90.34\textsubscript{$\pm$0.82} & -0.66 \\
			Rest16 & ACBS-only    & 82.25\textsubscript{$\pm$1.43} & 6.49 & 94.11\textsubscript{$\pm$0.59} & 3.11 \\
			Rest16 & Full         & \textbf{84.38\textsubscript{$\pm$1.65}} & 8.62 & \textbf{94.87\textsubscript{$\pm$0.25}} & 3.87 \\
			\bottomrule
		\end{tabular}
	}
\end{table}

\begin{table*}[t]
	\small
	\centering
	\caption{Performance comparison across different backbone encoders (DeBERTa, RoBERTa, BERT).}
	\label{tab:model_agnostic_comparison}
    \setlength{\tabcolsep}{3pt}
	\resizebox{\textwidth}{!}{%
		\begin{tabular}{llcccccccccccc}
			\toprule
			\multirow{2}{*}{\textbf{Dataset}} & \multirow{2}{*}{\textbf{Config}} &
			\multicolumn{4}{c}{\textbf{DeBERTa-base}} & \multicolumn{4}{c}{\textbf{RoBERTa-base}} & \multicolumn{4}{c}{\textbf{BERT-base}} \\
			\cmidrule(lr){3-6}\cmidrule(lr){7-10}\cmidrule(lr){11-14}
			& & \textbf{MF1} & \textbf{$\Delta$ MF1} & \textbf{Acc} & \textbf{$\Delta$ Acc} &
			\textbf{MF1} & \textbf{$\Delta$ MF1} & \textbf{Acc} & \textbf{$\Delta$ Acc} &
			\textbf{MF1} & \textbf{$\Delta$ MF1} & \textbf{Acc} & \textbf{$\Delta$ Acc} \\
			\midrule
			Lap14 & Encoder-only & 75.03 & 0.00 & 79.16 & 0.00 & 76.85 & 0.00 & 80.00 & 0.00 & 73.88 & 0.00 & 78.43 & 0.00 \\
			Lap14 & DORA-only    & 75.72 & 0.69 & 79.63 & 0.47 & 77.09 & 0.24 & 81.42 & 1.42 & 74.34 & 0.46 & 77.95 & -0.48 \\
			Lap14 & ACBS-only    & 79.97 & 4.94 & 82.62 & 3.46 & 79.93 & 3.08 & 82.99 & 2.99 & 75.94 & 2.06 & 79.53 & 1.10 \\
			Lap14 & DABS (Full)  & 81.56 & 6.53 & 84.41 & 5.25 & 80.84 & 3.99 & 83.62 & 3.62 & 76.94 & 3.06 & 80.31 & 1.88 \\
			\midrule
			Res14 & Encoder-only & 76.33 & 0.00 & 84.48 & 0.00 & 72.70 & 0.00 & 81.83 & 0.00 & 70.61 & 0.00 & 80.21 & 0.00 \\
			Res14 & DORA-only    & 73.18 & -3.15 & 83.02 & -1.46 & 74.37 & 1.67 & 83.17 & 1.34 & 71.56 & 0.95 & 80.30 & 0.09 \\
			Res14 & ACBS-only    & 83.15 & 6.82 & 88.81 & 4.33 & 80.93 & 8.23 & 87.02 & 5.19 & 77.93 & 7.32 & 84.51 & 4.30 \\
			Res14 & DABS (Full)  & 84.87 & 8.54 & 89.76 & 5.28 & 81.09 & 8.39 & 87.29 & 5.46 & 80.47 & 9.86 & 86.12 & 5.91 \\
			\midrule
			Res15 & Encoder-only & 71.39 & 0.00 & 85.85 & 0.00 & 67.73 & 0.00 & 83.03 & 0.00 & 65.74 & 0.00 & 82.10 & 0.00 \\
			Res15 & DORA-only    & 66.60 & -4.79 & 84.56 & -1.29 & 66.82 & -0.91 & 83.21 & 0.18 & 63.38 & -2.36 & 81.92 & -0.18 \\
			Res15 & ACBS-only    & 72.60 & 1.21 & 88.19 & 2.34 & 70.55 & 2.82 & 87.27 & 4.24 & 63.14 & -2.60 & 83.03 & 0.93 \\
			Res15 & DABS (Full)  & 74.06 & 2.67 & 89.18 & 3.33 & 71.37 & 3.64 & 88.01 & 4.98 & 66.98 & 1.24 & 83.95 & 1.85 \\
			\midrule
			Res16 & Encoder-only & 75.76 & 0.00 & 91.00 & 0.00 & 73.79 & 0.00 & 90.51 & 0.00 & 70.98 & 0.00 & 88.87 & 0.00 \\
			Res16 & DORA-only    & 72.82 & -2.94 & 90.34 & -0.66 & 73.91 & 0.12 & 89.85 & -0.66 & 68.81 & -2.17 & 88.54 & -0.33 \\
			Res16 & ACBS-only    & 82.25 & 6.49 & 94.11 & 3.11 & 81.74 & 7.95 & 93.62 & 3.11 & 72.84 & 1.86 & 90.02 & 1.15 \\
			Res16 & DABS (Full)  & 84.38 & 8.62 & 94.87 & 3.87 & 82.26 & 8.47 & 94.11 & 3.60 & 78.79 & 7.81 & 92.47 & 3.60 \\
			\bottomrule
		\end{tabular}%
	}
\end{table*}

\subsection{Cross-dataset Consistency and Discriminability}

Table~\ref{tab:sa1_effectiveness_extended} decomposes DABS into its two functional parts: (i) DORA, which constructs an ordered depth substrate intended for reuse and depth-sensitive querying, and (ii) ACBS, which performs aspect-conditioned retrieval over tokens and depths.

Two patterns are consistent across datasets. First, \textbf{ACBS-only} provides the majority of the gains over the encoder-only baseline, indicating that aspect-conditioned evidence localization and depth selection are the primary drivers of discriminability. Second, \textbf{DORA-only} is not designed to be a standalone classifier. Its role is to restructure representations for reuse. In other words, it is intentionally aspect-agnostic, and its role is to reshape representations into a reusable, depth-indexed substrate rather than to optimize aspect-specific decision boundaries. When combined, DORA enables ACBS to query a structured depth bank with minimal per-aspect cost, yielding the strongest and most stable improvements in the full model.

\subsection{Backbone-Agnostic Robustness}
\label{app:backbone_agnostic}

To verify that DABS is not tied to a specific pretrained encoder, we evaluate the same architecture under alternative backbone encoders, replacing DeBERTa-base with RoBERTa-base and BERT-base while keeping DORA/ACBS unchanged.
We follow the same training protocol as in the main experiments (same datasets, optimization, and three random seeds), and report MF1/Acc together with improvements over the corresponding encoder-only baseline.

Table~\ref{tab:model_agnostic_comparison} shows that DABS yields consistent gains across backbones and datasets.
For instance, DABS improves MF1 on Rest14 by +8.39 (RoBERTa-base ~\cite{liu-eyal-2019-roberta}) and +9.86 (BERT-base ~\cite{devlin-etal-2019-bert}), and on Rest16 by +8.47 (RoBERTa-base) and +7.81 (BERT-base).
Similar improvements are observed on Lap14 and Rest15, indicating that our single-pass, depth-selective querying mechanism is backbone-agnostic. Overall, these results support that the benefits of reusable depth substrates and aspect-conditioned depth allocation arise from the proposed readout formulation rather than encoder-specific artifacts.

\begin{table*}[t]
	\centering
	\small
	\caption{Single-aspect ($M{=}1$) efficiency breakdown: FLOPs, throughput, latency (p50/p95), parameter counts, and relative change vs.\ encoder-only. \textbf{Bold} marks the best value per column.}
	\label{tab:sa2_efficiency_extended}
	\setlength{\tabcolsep}{2.5pt}
	\begin{tabular}{llcc|ccc|cc|ccc}
		\toprule
		& & & & \multicolumn{3}{c|}{Latency (ms)} & \multicolumn{2}{c|}{Parameters} & \multicolumn{3}{c}{$\Delta$ vs Encoder (\%)} \\
		Dataset & Config & FLOPs & queries/s & p50 & p95 & p95/p50 & Total & Added & FLOPs & p50 & queries/s \\
		\midrule
		Lap14 & Encoder-only & \textbf{14.48} & \textbf{114.2} & \textbf{7.6} & 15.6 & 2.05 & \textbf{184.4} & \textbf{1.2} & \textbf{0.0} & \textbf{0.0} & \textbf{0.0} \\
		Lap14 & DORA-only    & 15.48 & 16.8  & 16.8 & 133.8 & 7.96 & 193.0 & 9.8 & 6.9 & 121.1 & -85.3 \\
		Lap14 & ACBS-only    & 14.58 & 69.9  & 9.0  & \textbf{10.0} & \textbf{1.11} & 192.1 & 8.9 & 0.7 & 18.4 & -38.8 \\
		Lap14 & Full         & 15.58 & 23.5  & 35.0 & 135.1 & 3.86 & 196.8 & 13.6 & 7.6 & 360.5 & -79.4 \\
		\midrule
		Rest14 & Encoder-only & \textbf{13.88} & 42.8 & \textbf{8.4} & 56.5 & 6.73 & \textbf{184.4} & \textbf{1.2} & \textbf{0.0} & 0.0 & 0.0 \\
		Rest14 & DORA-only    & 14.81 & 37.6 & 22.2 & 34.9 & 1.57 & 193.0 & 9.8 & 6.7 & 164.3 & -12.1 \\
		Rest14 & ACBS-only    & 13.98 & \textbf{116.9} & \textbf{8.4} & \textbf{9.1} & \textbf{1.08} & 192.1 & 8.9 & 0.7 & \textbf{0.0} & \textbf{173.1} \\
		Rest14 & Full         & 14.90 & 58.3 & 17.6 & 20.4 & 1.16 & 196.8 & 13.6 & 7.3 & 109.5 & 36.2 \\
		\midrule
		Rest15 & Encoder-only & \textbf{13.37} & \textbf{126.1} & \textbf{7.4} & \textbf{8.2} & 1.11 & \textbf{184.4} & \textbf{1.2} & \textbf{0.0} & \textbf{0.0} & \textbf{0.0} \\
		Rest15 & DORA-only    & 14.23 & 39.8 & 19.0 & 36.7 & 1.93 & 193.0 & 9.8 & 6.4 & 156.8 & -68.4 \\
		Rest15 & ACBS-only    & 13.46 & 116.7 & 8.5 & 8.9 & \textbf{1.05} & 192.1 & 8.9 & 0.7 & 14.9 & -7.5 \\
		Rest15 & Full         & 14.31 & 54.6 & 18.0 & 22.1 & 1.23 & 196.8 & 13.6 & 7.0 & 143.2 & -56.7 \\
		\midrule
		Rest16 & Encoder-only & \textbf{14.48} & \textbf{107.3} & \textbf{8.1} & \textbf{14.1} & 1.74 & \textbf{184.4} & \textbf{1.2} & \textbf{0.0} & \textbf{0.0} & \textbf{0.0} \\
		Rest16 & DORA-only    & 15.48 & 52.5 & 18.2 & 22.3 & 1.23 & 193.0 & 9.8 & 6.9 & 124.7 & -51.1 \\
		Rest16 & ACBS-only    & 14.58 & 68.5 & 14.3 & 16.4 & \textbf{1.15} & 192.1 & 8.9 & 0.7 & 76.5 & -36.2 \\
		Rest16 & Full         & 15.58 & 24.3 & 23.3 & 131.9 & 5.66 & 196.8 & 13.6 & 7.6 & 187.7 & -77.4 \\
		\bottomrule
	\end{tabular}
\end{table*}
\begin{table}[t]
	\centering
    \small
	\caption{Seed-wise MF1/Acc for paired tests (Full vs.\ Enc) under 3 seeds. $\Delta$ is reported in percentage points (pp).}
	\label{tab:sa7_seed_alignment}
	\setlength{\tabcolsep}{3pt}
    \begin{tabular}{lc|ccc|ccc}
        \toprule
        \multirow{2}{*}{Dataset} & \multirow{2}{*}{Seed} &
        \multicolumn{3}{c|}{MF1} &
        \multicolumn{3}{c}{Acc} \\
        & & Full & Enc & $\Delta$ (pp) & Full & Enc & $\Delta$ (pp) \\
        \midrule
        Lap14  & 42  & 81.22 & 75.36 & 5.86 & 84.25 & 79.21 & 5.04 \\
        Lap14  & 123 & 81.73 & 75.18 & 6.55 & 84.41 & 79.37 & 5.04 \\
        Lap14  & 456 & 81.71 & 74.55 & 7.16 & 84.57 & 78.90 & 5.67 \\
        \midrule
        Rest14 & 42  & 84.57 & 76.43 & 8.14 & 89.70 & 84.24 & 5.46 \\
        Rest14 & 123 & 85.42 & 77.52 & 7.90 & 89.97 & 85.05 & 4.92 \\
        Rest14 & 456 & 84.63 & 75.05 & 9.58 & 89.62 & 84.15 & 5.47 \\
        \midrule
        Rest15 & 42  & 73.39 & 70.10 & 3.29 & 89.11 & 85.42 & 3.69 \\
        Rest15 & 123 & 75.90 & 72.10 & 3.80 & 90.41 & 86.90 & 3.51 \\
        Rest15 & 456 & 72.89 & 71.97 & 0.92 & 88.01 & 85.24 & 2.77 \\
        \midrule
        Rest16 & 42  & 86.11 & 78.35 & 7.76 & 95.09 & 91.49 & 3.60 \\
        Rest16 & 123 & 82.83 & 73.37 & 9.46 & 94.60 & 90.83 & 3.77 \\
        Rest16 & 456 & 84.19 & 75.57 & 8.62 & 94.93 & 90.67 & 4.26 \\
        \bottomrule
    \end{tabular}
\end{table}

\subsection{Overall Efficiency and Resource Overhead}

Table~\ref{tab:sa2_efficiency_extended} provides a component-wise efficiency breakdown under a single-aspect protocol ($M{=}1$). Each measurement corresponds to one \emph{(sentence, aspect)} query and therefore does not benefit from cross-aspect reuse. Under this setting, the full model can be slower than Encoder-only because it pays a fixed overhead to construct and retain the depth substrate (primarily from DORA). This table is included to isolate that fixed cost at $M{=}1$. For the deployment-relevant multi-aspect setting ($M\ge2$), DABS computes the encoder+\textsc{DORA} substrate once per sentence and reuses it across aspects, yielding the amortized gains shown in Figure~\ref{fig:multi_aspect} (in the main paper).

\paragraph{Why single-aspect can be slower.}
We include the $M{=}1$ setting solely to isolate fixed construction costs, not as a target deployment scenario. Under $M{=}1$, DABS cannot amortize the fixed cost of substrate construction and therefore can be slower than encoder-only. This is expected as DORA introduces additional kernels (local convolutions) and depth-ordered integration (DepthGRU), and ACBS adds lightweight MLP-based gating. We report p50/p95 to separate typical from tail behavior whereby large p95/p50 ratios often reflect sensitivity to system-level variance (\textit{e.g.}, cache state and kernel scheduling) rather than algorithmic instability. The deployment-relevant takeaway is thus not the $M{=}1$ regime, but the multi-aspect regime where encoder+DORA is computed once per sentence and reused across aspects (Figure~\ref{fig:multi_aspect} in the main paper).

\section{Significance Tests and Seed-level Scores}
\label{app:significance}

Table~\ref{tab:sa7_seed_alignment} reports per-seed MF1/Acc for DABS (Full) and the encoder-only baseline. Across all four datasets, the seed-wise differences were predominantly positive, aligning with the aggregate improvements reported in Table~\ref{tab:main_results} (in the main paper). These aligned scores were used for paired statistical tests in the main paper.

\begin{table*}[t]
	\centering
	\caption{Per-seed multilingual results (Acc/MF1) across 3 seeds for French, Russian, and Spanish.}
	\label{tab:multilingual_detailed}
    \setlength{\tabcolsep}{3pt}
	\resizebox{\textwidth}{!}{%
		\begin{tabular}{l | ccc | ccc | ccc | ccc | ccc | ccc }
			\toprule
			& \multicolumn{6}{c|}{\textbf{French}} & \multicolumn{6}{c|}{\textbf{Russian}} & \multicolumn{6}{c}{\textbf{Spanish}} \\
			\cmidrule(lr){2-7} \cmidrule(lr){8-13} \cmidrule(lr){14-19}
			& \multicolumn{3}{c|}{Acc (\%)} & \multicolumn{3}{c|}{MF1 (\%)} & \multicolumn{3}{c|}{Acc (\%)} & \multicolumn{3}{c|}{MF1 (\%)} & \multicolumn{3}{c|}{Acc (\%)} & \multicolumn{3}{c}{MF1 (\%)} \\
			\cmidrule(lr){2-4} \cmidrule(lr){5-7} \cmidrule(lr){8-10} \cmidrule(lr){11-13} \cmidrule(lr){14-16} \cmidrule(lr){17-19}
			\textbf{Config (seed)} & 42 & 123 & 456 & 42 & 123 & 456 & 42 & 123 & 456 & 42 & 123 & 456 & 42 & 123 & 456 & 42 & 123 & 456 \\
			\midrule
			Baseline     & 85.08 & 86.31 & 85.54 & 74.01 & 74.73 & 75.27 & 83.93 & 84.04 & 85.52 & 72.76 & 73.06 & 74.06 & 88.86 & 88.43 & 89.56 & 72.05 & 69.65 & 73.50 \\
			DORA-only    & 84.77 & 85.38 & 84.92 & 74.11 & 74.00 & 73.80 & 85.10 & 84.46 & 85.10 & 73.16 & 72.56 & 72.88 & 87.73 & 89.28 & 87.87 & 68.42 & 73.28 & 72.04 \\
			ACBS-only    & 89.54 & 88.31 & 87.85 & 80.97 & 77.65 & 78.88 & 88.69 & 87.95 & 89.22 & 78.22 & 75.74 & 79.29 & 92.95 & 91.96 & 93.23 & 78.18 & 76.78 & 78.95 \\
			Full         & 90.31 & 88.77 & 88.77 & 82.18 & 79.51 & 81.12 & 89.11 & 89.75 & 89.32 & 79.95 & 79.99 & 79.72 & 94.08 & 92.95 & 92.38 & 81.70 & 79.39 & 79.07 \\
			\bottomrule
	\end{tabular}}
\end{table*}
\begin{table*}[t]
	\centering
	\small
	\caption{Depth-region masking on multilingual benchmarks using best checkpoints. Each condition retains one 2-layer band within the last $K{=}6$ layers. $\Delta$ is the best-worst MF1 gap (\%).}
	\label{tab:depth-regions}
    \setlength{\tabcolsep}{3pt}
    \begin{tabular}{llcccccc}
        \toprule
        Lang & Config & Base MF1 & Shallow (L\textsubscript{7-8}) & Middle (L\textsubscript{9-10}) & Deep (L\textsubscript{11-12}) & $\Delta$(Best-Worst) & Best Region \\
        \midrule
        Spanish & ACBS-only & 78.95 & 78.82 & 79.21 & 79.85 & 1.03 & Deep \\
        Spanish & Full      & 81.70 & 81.29 & 81.37 & 81.70 & 0.41 & Deep \\
        French & ACBS-only & 80.97 & 80.97 & 80.94 & 80.05 & 0.92 & Shallow \\
        French & Full      & 82.18 & 82.32 & 81.67 & 81.42 & 0.90 & Shallow \\
        Russian & ACBS-only & 79.29 & 78.77 & 78.78 & 79.40 & 0.63 & Deep \\
        Russian & Full      & 79.99 & 80.23 & 80.02 & 80.73 & 0.71 & Deep \\
        \bottomrule
    \end{tabular}%
\end{table*}

\begin{table*}[t]
	\centering
	\small
	\caption{Single-layer depth controls on multilingual benchmarks using best checkpoints. Rand-2L keeps a random contiguous 2-layer block (20 trials) and $\Delta$ are relative to Base.}	
	\label{tab:depth-controls-single}
    \setlength{\tabcolsep}{3pt}
	\resizebox{\textwidth}{!}{%
		\begin{tabular}{llcccc}
			\toprule
			Lang & Config & Base (Acc/MF1) & Rand-2L MF1 ($\mu\pm\sigma$, $\Delta$) & Best-1L (layer, MF1, $\Delta$) & Worst-1L (layer, MF1, $\Delta$) \\
			\midrule
			Spanish & ACBS-only & 93.23/78.95 & 79.22$\pm$0.37 ($\greenup$0.27) & d4: 79.52 ($\greenup$0.57) & d1: 78.64 ($\reddown$0.31) \\
			Spanish & Full      & 94.08/81.70 & 81.61$\pm$0.16 ($\reddown$0.09) & d4: 82.17 ($\greenup$0.47) & d6: 81.29 ($\reddown$0.41) \\
			French & ACBS-only & 89.54/80.97 & 80.63$\pm$0.38 ($\reddown$0.34) & d1: 81.73 ($\greenup$0.76) & d3: 79.54 ($\reddown$1.43) \\
			French & Full      & 90.31/82.18 & 81.79$\pm$0.30 ($\reddown$0.39) & d2: 82.55 ($\greenup$0.37) & d5: 81.42 ($\reddown$0.76) \\
			Russian & ACBS-only & 89.22/79.29 & 78.96$\pm$0.27 ($\reddown$0.33) & d6: 79.33 ($\greenup$0.04) & d3: 78.65 ($\reddown$0.64) \\
			Russian & Full      & 89.75/79.99 & 80.19$\pm$0.19 ($\greenup$0.20) & d1: 80.65 ($\greenup$0.66) & d3: 79.84 ($\reddown$0.15) \\
			\bottomrule
		\end{tabular}%
	}
\end{table*}

\section{Additional Multilingual Analyses}
\label{app:multilingual_details}

\subsection{Per-seed Multilingual Results}

Table~\ref{tab:multilingual_detailed} lists per-seed results for all configurations on French, Russian, and Spanish.
ACBS-only delivered the strongest single-component improvements across languages, while the full model was the most stable overall. This pattern suggests that the aspect-conditioned readout in ACBS transfers well cross-lingually, whereas DORA primarily functions as a reusable depth substrate that supports single-pass reuse and depth-ordered querying in the full system.

\subsection{Depth Controls in the Multilingual Setting}
\label{app:depth_controls}

We evaluated whether depth allocation behaved as structured specialization rather than incidental layer mixing by applying inference-time depth controls within the last $K{=}6$ layers. These controls probe whether performance is sensitive to \emph{which} depth band is available, rather than benefiting from arbitrary layer mixing. If depth allocation is functionally meaningful, restricting access to certain bands should change MF1 in a structured way. We observe that the preferred depth region can vary across languages (Table~\ref{tab:depth-regions}), which is plausible under domain and distribution shifts. Rand-2L in Table~\ref{tab:depth-controls-single} serves as a sanity-check baseline: it quantifies the expected variability when keeping a random contiguous band, making best/worst single-layer effects easier to interpret.

\subsubsection{Region Masking}

Table~\ref{tab:depth-regions} reports MF1 under shallow/middle/deep region masking, where each condition retained a contiguous 2-layer band within the last $K$ layers. The best-worst gap $\Delta$ indicates that performance depended on which depth band remained, and that the preferred band varied by language and configuration.

\subsubsection{Single-layer Controls}

Table~\ref{tab:depth-controls-single} further breaks this down by retaining a single layer at a time, and compares against a randomized contiguous 2-layer baseline (Rand-2L). The consistent best/worst layer differences provide additional evidence that certain layers serve as stronger evidence sources than others, supporting the functional relevance of depth allocation under multilingual transfer.

\section{Additional Ablations}
\label{app:add_ablation}

\subsection{Accuracy Ablations}
\label{app:ablation_acc_app}

Table~\ref{tab:ablation_acc} reports $\mathrm{Acc}_{\text{mean}}$ under the same ablation settings as the MF1 ablations in Table~\ref{tab:ablation_mf1}. The accuracy trends closely mirrored MF1: removing ACBS components (especially token selection and gated fusion) produced the most consistent degradation, while removing DORA components led to smaller but still noticeable drops. The alignment across metrics suggests that the gains are not artifacts of a particular evaluation measure.

\begin{table}[t]
	\small
	\centering
	\caption{Ablations on $\mathrm{Acc}_{\text{mean}}$. Values are reported as Acc$^{\Delta}$, where $\Delta$ is the change (pp) from the full model.}	
	\label{tab:ablation_acc}
	\setlength{\tabcolsep}{2pt}
	\resizebox{\linewidth}{!}{%
		\begin{tabular}{l|cccc|c}
			\toprule
			\multirow{2}{*}{Config} & Lap14 & Rest14 & Rest15 & Rest16 & Avg \\
			& Acc$^{\Delta}$ & Acc$^{\Delta}$ & Acc$^{\Delta}$ & Acc$^{\Delta}$ & $\Delta$ \\
			\midrule
			\textbf{DABS (Full)} & \textbf{84.41}$^{\scriptsize 0.00}$ & \textbf{89.76}$^{\scriptsize 0.00}$ & \textbf{89.18}$^{\scriptsize 0.00}$ & \textbf{94.87}$^{\scriptsize 0.00}$ & 0.00 \\
			\midrule
			\textit{w/o Regularizers} & & & & & \\
			~- Sparsity      & 82.94$^{\scriptsize \reddown1.47}$ & 89.05$^{\scriptsize \reddown0.71}$ & 88.50$^{\scriptsize \reddown0.68}$ & 94.33$^{\scriptsize \reddown0.54}$ & $\reddown$0.85 \\
			~- Span Masking  & 82.62$^{\scriptsize \reddown1.79}$ & 88.72$^{\scriptsize \reddown1.04}$ & 88.87$^{\scriptsize \reddown0.31}$ & 94.76$^{\scriptsize \reddown0.11}$ & $\reddown$0.81 \\
			~- Gate Entropy  & 82.52$^{\scriptsize \reddown1.89}$ & 89.49$^{\scriptsize \reddown0.27}$ & 89.18$^{\scriptsize 0.00}$  & 94.33$^{\scriptsize \reddown0.54}$ & $\reddown$0.68 \\
			\midrule
			\textit{w/o DORA} & & & & & \\
			~- DepthGRU       & 83.57$^{\scriptsize \reddown0.84}$ & 89.05$^{\scriptsize \reddown0.71}$ & 88.75$^{\scriptsize \reddown0.43}$ & 94.16$^{\scriptsize \reddown0.71}$ & $\reddown$0.67 \\
			~- LCP (Pooling)  & 82.83$^{\scriptsize \reddown1.58}$ & 88.63$^{\scriptsize \reddown1.13}$ & 88.50$^{\scriptsize \reddown0.68}$ & 93.84$^{\scriptsize \reddown1.03}$ & $\reddown$1.11 \\
			\midrule
			\textit{w/o ACBS} & & & & & \\
			~- Token Sel.     & 82.05$^{\scriptsize \reddown2.36}$ & 88.84$^{\scriptsize \reddown0.92}$ & 88.01$^{\scriptsize \reddown1.17}$ & 93.94$^{\scriptsize \reddown0.93}$ & $\reddown$1.35 \\
			~- Layer Sel.     & 83.31$^{\scriptsize \reddown1.10}$ & 88.27$^{\scriptsize \reddown1.49}$ & 87.95$^{\scriptsize \reddown1.23}$ & 94.38$^{\scriptsize \reddown0.49}$ & $\reddown$1.08 \\
			~- Gated Fusion   & 82.26$^{\scriptsize \reddown2.15}$ & 87.94$^{\scriptsize \reddown1.82}$ & 88.56$^{\scriptsize \reddown0.62}$ & 94.33$^{\scriptsize \reddown0.54}$ & $\reddown$1.28 \\
			\bottomrule
		\end{tabular}%
	}
\end{table}

\section{Cross-Dataset Stress-Test Splits and Depth-Order Controls}
\label{app:cross_dataset_controls}

This appendix reports controlled analyses on three cross-dataset stress-test splits designed to emphasize sentence-level difficulty patterns. The goal is to test whether the ordered depth integration in DORA behaves as a functional mechanism, rather than a cosmetic architectural choice.

\subsection{Split Construction}
\label{app:cross_dataset_construction}

We pooled training splits from SemEval datasets (Lap14, Rest14, Rest15, Rest16) into a single training set, and pooled their test splits into a single test set. From each pool, we constructed three stress-test subsets using explicit filtering rules:

\begin{itemize}[leftmargin=*, itemsep=0pt, topsep=2pt]
	\item \textbf{Long Sentences.} We computed subword length using the DeBERTa tokenizer and retained sentences at or above the P90 length threshold of the corresponding pool.
	\item \textbf{Multi-Aspect Conflict.} We retained sentences containing at least one positive aspect and at least one negative aspect within the same sentence.
	\item \textbf{Complex Negation.} We retained sentences that contained a negation cue (\textit{e.g.}, \emph{no}, \emph{not}, \emph{never}, \emph{n't}, \emph{without}) and had subword length greater than 40.
\end{itemize}

Unless stated otherwise, results in this appendix used $K{=}6$ and report mean and standard deviation over $N{=}3$ seeds.

Table~\ref{tab:cross_dataset_statistics} reports class distributions for the three stress-test splits.
Counts are aspect-level instances. The Multi-Aspect Conflict split contained very few neutral instances, which can increase variance in macro-F1.

\begin{table}[t]
	\centering
	\caption{Class distributions for cross-dataset stress-test splits. Counts are aspect-level instances in train/test.}	
	\label{tab:cross_dataset_statistics}
    \setlength{\tabcolsep}{3pt}
	\resizebox{\linewidth}{!}{%
		\begin{tabular}{lcccccc}
			\toprule
			\textbf{Dataset} & \multicolumn{2}{c}{\textbf{Positive}} &
			\multicolumn{2}{c}{\textbf{Neutral}} & \multicolumn{2}{c}{\textbf{Negative}} \\
			\cmidrule(lr){2-3}\cmidrule(lr){4-5}\cmidrule(lr){6-7}
			& \textbf{Train} & \textbf{Test} & \textbf{Train} & \textbf{Test} &
			\textbf{Train} & \textbf{Test} \\
			\midrule
			Long Sentences & 677 & 274 & 282 & 78 & 538 & 156 \\
			Multi-Aspect Conflict & 372 & 96  & 38  & 6  & 368 & 96  \\
			Complex Negation & 116 & 29 & 44 & 8 & 154 & 40 \\
			\bottomrule
		\end{tabular}%
	}
\end{table}\begin{table}[t]
	\centering
	\caption{Layer-order ablation on cross-dataset stress-test splits ($K{=}6$). Results are mean$\pm$std over 3 seeds. $\Delta$MF1 is relative to the normal order (pp).}	
	\label{tab:layer_order_ablation_k6}
	\setlength{\tabcolsep}{3pt}
	\resizebox{\linewidth}{!}{%
		\begin{tabular}{llccc}
			\toprule
			\textbf{Dataset} & \textbf{Order} & \textbf{Acc} & \textbf{MF1} & $\Delta$\textbf{MF1} \\
			\midrule
			\multirow{3}{*}{\parbox{2.1cm}{Long \\ Sentences}}
			& normal   & 82.74\textsubscript{$\pm$ 1.08} & 77.97\textsubscript{$\pm$ 0.72} & 0.00 \\
			& reversed & 82.26\textsubscript{$\pm$ 0.60} & 76.44\textsubscript{$\pm$ 0.21} & -1.53 \\
			& shuffled & 82.19\textsubscript{$\pm$ 1.46} & 77.22\textsubscript{$\pm$ 1.39} & -0.75 \\
			\midrule
			\multirow{3}{*}{\parbox{2.1cm}{Multi-Aspect \\ Conflict}}
			& normal   & 95.40\textsubscript{$\pm$ 1.15} & 76.88\textsubscript{$\pm$ 10.80} & 0.00 \\
			& reversed & 95.79\textsubscript{$\pm$ 0.88} & 74.10\textsubscript{$\pm$  9.55} & -2.78 \\
			& shuffled & 95.40\textsubscript{$\pm$ 1.15} & 74.28\textsubscript{$\pm$  9.77} & -2.60 \\
			\midrule
			\multirow{3}{*}{\parbox{2.1cm}{Complex \\ Negation}}
			& normal   & 83.81\textsubscript{$\pm$ 2.18} & 80.62\textsubscript{$\pm$ 1.49} & 0.00 \\
			& reversed & 83.33\textsubscript{$\pm$ 2.97} & 79.14\textsubscript{$\pm$ 1.19} & -1.48 \\
			& shuffled & 83.81\textsubscript{$\pm$ 2.18} & 80.10\textsubscript{$\pm$ 1.55} & -0.52 \\
			\bottomrule
	\end{tabular}}
\end{table}

\begin{table}[t]
	\centering
	\caption{DepthGRU ablation on cross-dataset stress-test splits ($K{=}6$). Results are mean$\pm$std over 3 seeds (Acc/MF1, \%).}	
	\label{tab:depth_gru_ablation_k6}
	\setlength{\tabcolsep}{3pt}
	\resizebox{\linewidth}{!}{%
		\begin{tabular}{llcc}
			\toprule
			\textbf{Dataset} & \textbf{Setting} & \textbf{Acc} & \textbf{MF1} \\
			\midrule
			\multirow{2}{*}{\parbox{2.1cm}{Long \\ Sentences}}
			& w DepthGRU & 82.95\textsubscript{$\pm$ 0.91} & 78.44\textsubscript{$\pm$ 0.76} \\
			& w/o DepthGRU & 82.61\textsubscript{$\pm$ 1.57} & 76.93\textsubscript{$\pm$ 2.14} \\
			\midrule
			\multirow{2}{*}{\parbox{2.1cm}{Multi-Aspect \\ Conflict}}
			& w DepthGRU & 95.40\textsubscript{$\pm$ 1.15} & 76.88\textsubscript{$\pm$ 10.80} \\
			& w/o DepthGRU & 94.83\textsubscript{$\pm$ 0.57} & 71.32\textsubscript{$\pm$  6.29} \\
			\midrule
			\multirow{2}{*}{\parbox{2.1cm}{Complex \\ Negation}}
			& w DepthGRU  & 83.81\textsubscript{$\pm$ 2.18} & 80.62\textsubscript{$\pm$ 1.49} \\
			& w/o DepthGRU & 82.86\textsubscript{$\pm$ 2.86} & 77.69\textsubscript{$\pm$ 3.36} \\
			\bottomrule
	\end{tabular}}
\end{table}

\subsection{Layer-Order Ablation ($K{=}6$)}
\label{app:layer_order_ablation}

DORA integrates the last $K$ layers using a depth-ordered recurrence. If the depth order is meaningful, disrupting that order should reduce performance under stress-test conditions. We used a DeBERTa-base backbone for this controlled study. We evaluated three conditions: (i) normal order, (ii) reversed order, and (iii) shuffled order. Table~\ref{tab:layer_order_ablation_k6} reports mean$\pm$std over three seeds, with $\Delta$MF1 computed relative to the normal order within each split.

DepthGRU is explicitly \emph{depth-ordered}. It accumulates representations from shallower to deeper layers, encouraging monotonic refinement from lexical/local cues to more abstract compositional features. Reversing the order forces the recurrence to integrate abstractions before grounding signals, which can blur compositional dependencies and reduce robustness under long-context or negation-heavy conditions. Shuffling disrupts the monotonicity less systematically than full reversal, which is consistent with the smaller average degradation observed under the shuffled condition.

\subsection{DepthGRU Ablation}
\label{app:depth_gru_ablation}

This ablation isolates the contribution of DepthGRU inside DORA by removing the recurrence while keeping the remaining components unchanged. We used a DeBERTa-base backbone with $K{=}6$ and $N{=}3$ seeds.
Table~\ref{tab:depth_gru_ablation_k6} compares the full model (Base) against the variant without DepthGRU (w/o DepthGRU).

Removing DepthGRU reduced macro-F1 across all three stress-test splits, with the largest drop on Multi-Aspect Conflict and Complex Negation. Together with the layer-order ablation, these results support that ordered depth integration contributes materially under conditions that emphasize long context, polarity conflict, and negation-driven composition.

We observe that accuracy can remain relatively stable while macro-F1 drops, especially on conflict/negation splits.
This pattern suggests that DepthGRU mainly improves \emph{minority-class} and \emph{hard-case} discrimination (which MF1 is sensitive to), rather than only boosting majority predictions.
This is consistent with our hypothesis that ordered depth integration provides additional compositional capacity that matters most when superficial lexical cues are insufficient.

\section{AI Assistants}
AI assistants were used solely to support language editing and clarity, including minor rephrasing and grammatical corrections. All technical content, modeling decisions, experimental design, results, and interpretations were conceived, implemented, and verified by the authors. No AI system was used to generate experimental results, analyze data, or make scientific claims.

\end{document}